\documentclass[final]{clv2025}
\jvol{vv}
\jnum{nn}
\jyear{2025}
\usepackage{todonotes}

\usepackage{amsmath}
\usepackage{multirow}
\usepackage{booktabs}
\usepackage{enumitem}
\usepackage{amsfonts}
\usepackage{xcolor}
\usepackage{soul}
\usepackage{color, colortbl}
\usepackage{graphicx}
\usepackage{array}
\usepackage{caption}
\usepackage{amssymb}
\usepackage{arydshln} 
\usepackage{amsmath, amssymb, amsthm, enumerate, graphicx}
\usepackage{xcolor} 
\usepackage{url}
\usepackage{subcaption} 

\runningtitle{{\textit{Can LLMs reason over extended multilingual contexts?}}}
\runningauthor{Hengle, Bajpai, Dan, Chakraborty}

\definecolor{darkblue}{rgb}{0, 0, 0.5}
\definecolor{turquoise}{RGB}{83, 95, 189}
\definecolor{orange_acc2}{RGB}{242, 93, 13}
\definecolor{LightCyan}{rgb}{0.88,1,1}
\definecolor{yellow}{HTML}{FFB300}
\definecolor{blue}{HTML}{2196F3}
\definecolor{green}{HTML}{689F38}
\definecolor{red}{HTML}{EF5350}
\definecolor{ForestGreen}{RGB}{34,139,34}
\definecolor{navyblue}{rgb}{0.0, 0.0, 0.5} 

\newcommand{\amey}[1]{\textcolor{magenta}{\bf\small [todo --Amey]}}
\newcommand{\prasoon}[1]{\textcolor{orange}{\bf\small [todo --Prasoon]}}
\newcommand{\soham}[1]{\textcolor{purple}{\bf\small [todo --Soham]}}
\newcommand{\dataset}{\texttt{MLRBench}}

\begin{document}
\title{{\textit{Can LLMs reason over extended multilingual contexts?}} Towards long-context evaluation beyond retrieval and haystacks}

\author{Amey Hengle$^{1}$, Prasoon Bajpai$^{1}$, Soham Dan$^{2}$, Tanmoy Chakraborty\thanks{Corresponding Author: tanchak@iitd.ac.in}$^{1}$}

\affilblock{
    \affil{Indian Institute of Technology Delhi, New Delhi, India}
    \affil{Microsoft}
}

\maketitle

\begin{abstract}
Existing multilingual long-context benchmarks, often based on the popular needle-in-a-haystack test, primarily evaluate a model's ability to locate specific information buried within irrelevant texts. However, such a retrieval-centric approach is myopic and inherently limited, as successful recall alone does not indicate a model's capacity to reason over extended contexts. Moreover, these benchmarks are susceptible to data leakage, short-circuiting, and risk making the evaluation a priori identifiable. To address these limitations, we introduce \dataset, a new synthetic benchmark for multilingual long-context reasoning. Unlike existing benchmarks, \dataset\ goes beyond surface-level retrieval by including tasks that assess multi-hop inference, aggregation, and epistemic reasoning. Spanning seven languages, \dataset\ is designed to be parallel, resistant to leakage, and scalable to arbitrary context lengths. Our extensive experiments with an open-weight large language model (LLM) reveal a pronounced gap between high- and low-resource languages, particularly for tasks requiring the model to aggregate multiple facts or predict the absence of information. We also find that, in multilingual settings, LLMs effectively utilize less than 30\% of their claimed context length.  Although off-the-shelf Retrieval Augmented Generation helps alleviate this to a certain extent, it does not solve the long-context problem.  We open-source \dataset\ to enable future research in improved evaluation and training of multilingual LLMs. \footnote{The source code and dataset are available at \url{https://github.com/AmeyHengle/multilingual-long-context-reasoning}}
\end{abstract}

\begin{figure}[t]
\centering \includegraphics[width=0.75\columnwidth]{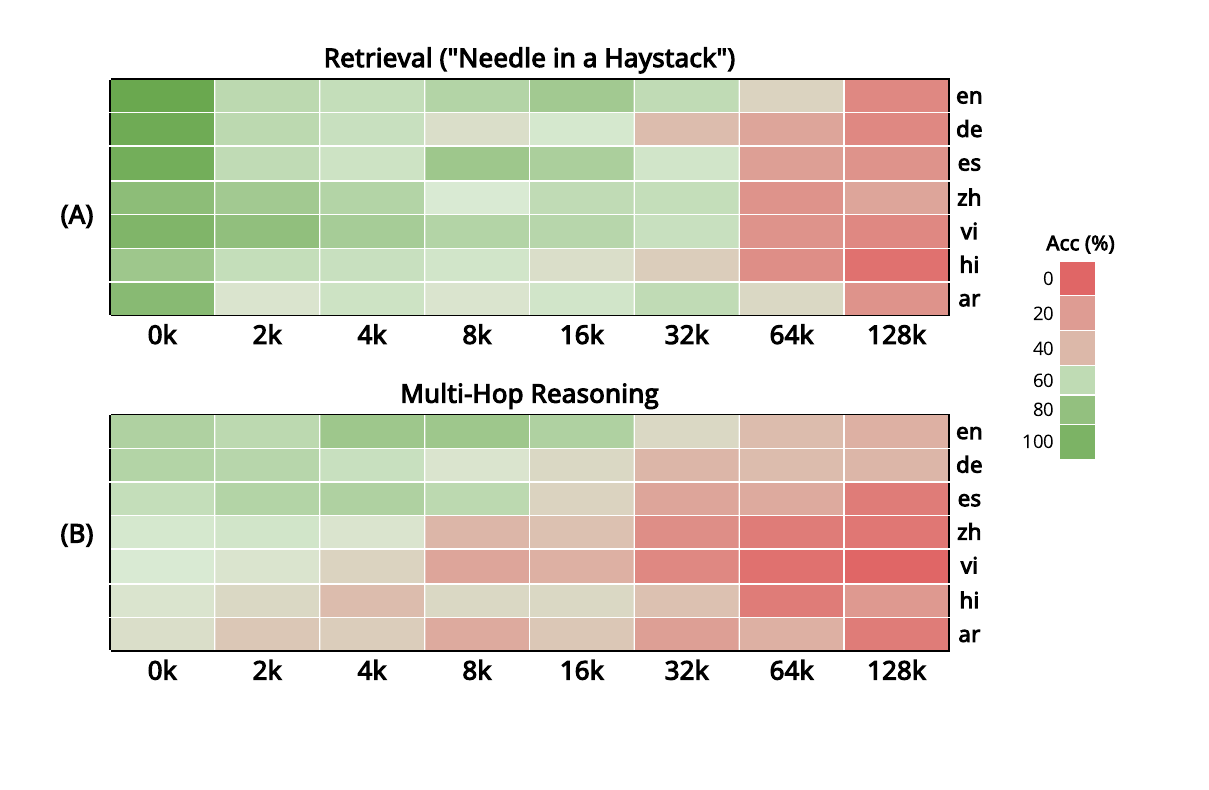}
\caption{Question-answering performance of {Llama-3.1-Instruct} on two tasks: (a) simple retrieval/recall versus (b) multi-hop reasoning. As context length increases, the performance drop in (b) is sharper than that in (a), especially for non-Latin languages. This underscores the difficulty of task (b) over task (a). Note that the tasks are (a) Basic Factoid QA and (b) Fact Chaining, respectively (refer to Section \ref{sec:task_categories}).} 
\label{fig:retrieval_vs_reasoning}
\vspace{-6mm}
\end{figure}
\section{Introduction}
\label{sec:introduction}

The landscape of Large Language Models (LLMs) has undergone a remarkable transformation in recent years, particularly in their ability to process long-input sequences. Just two years back, LLMs, including early versions of \text{GPT-3} and \text{Llama}, used to support a context window of no more than $4k$ tokens in both input and output \cite{brown2020languagemodelsfewshotlearners}. Fast forward to 2025, the context size of these models has expanded by several orders of magnitude, with newer LLMs like \text{Gemini-2.0} capable of processing up to a million tokens at once \cite{reid2024-gemini}. This dramatic increase in context window has unlocked new capabilities, allowing LLMs to infer from extremely long documents. As a result, their adoption and deployment have surged across a wide range of global-facing applications. For instance, applications involving dialogue systems, multi-document question answering, summarising long documents, or debugging lengthy code benefit from LLM's ability to retain and infer over-dispersed information within and across documents \cite{application-Lee_2022, application-rozière2024codellamaopenfoundation, application-shah2024multidocumentfinancialquestionanswering}.

Notwithstanding, some critical questions persist as we enter the long-context era of LLMs -- Can LLMs {fully utilize and reason over their claimed context window?} Do they {maintain comparable performance in multilingual settings, particularly for {agglutinative}, {morphologically rich} , and {low-resource} languages like Arabic or Hindi}, relative to English? Confronting these questions is vital,  as any attempts to increase LLM's context window are futile unless we achieve a conceptual and empirical understanding of their behavior in such settings.

\paragraph{\textbf{Drawbacks of existing evaluation frameworks}}
There is a growing research interest in evaluating the long-context behavior of LLMs, although most of it is predominantly limited to monolingual English settings \cite{bai-etal-Longbench-2024, Liu2023-LostITM, hsieh2024-Ruler}. A central focus of this research is to develop reliable benchmarks to assess how well LLMs handle extended contexts. Because realistic long-context evaluations are complex and resource-intensive \cite{karpinska2024-Thousand-and-one-pairs}, studies rely on synthetic benchmarks as \textit{proxies} for real-world performance \cite{hsieh2024-Ruler, bai-etal-Longbench-2024}. One such widely adopted proxy evaluation is the \textit{Needle-in-a-Haystack} (NIAH) test \cite{gregory2023-NIAH}, which frames the task as a retrieval problem -- measuring whether a model can locate specific facts within lengthy input contexts. Recent evaluation benchmarks are built on this framework by including multiple needles \cite{li2024needlebenchllmsretrievalreasoning, hsieh2024-Ruler} and question-answering \cite{bohnet2024longspanquestionansweringautomaticquestion, Liu2023-LostITM, zhang-infinite-bench-2024}. Although these benchmarks certainly add a good amount of depth, complexity, and realism to the task, they still adhere to the core evaluation principle of NIAH. Similarly, multilingual long-context benchmarks such as MLNeedle \cite{hengle2024_multilingualneedlehaystackinvestigating}, mLongRR \cite{mLongRR}, and ONERULER \cite{kim2025_rulermeasureallbenchmarking} extend this idea to languages other than English but still largely focus on evaluating \textit{retrieval} or \textit{recall} over long multilingual sequences \cite{michelangelo}. 

Recently, several studies have drawn attention to the design limitations of retrieval-centric NIAH tests, particularly their lack of effectiveness in evaluating LLM's reasoning capabilities \cite{goldman2024reallylongcontextneed, karpinska2024-Thousand-and-one-pairs}. The central argument is that a model’s ability to \textit{retrieve independent facts} from long contexts does not necessarily imply the ability to \textit{synthesize information} across those contexts \cite{michelangelo} In other words, successful retrieval alone should not be conflated with true comprehension.  Although LLMs may effectively extract isolated pieces of information, even from extended inputs, this does not guarantee they can follow logical connections, resolve contradictions, or maintain coherent reasoning over time \cite{karpinska2024thousandpairsnovelchallenge}. We empirically validate this distinction in Figure \ref{fig:retrieval_vs_reasoning}, where we compare model performance on retrieval- and reasoning-focused tasks under long-context settings. The results reveal a pronounced divergence: as the context length increases, the model maintains relatively stable performance on retrieval tasks but struggles significantly with reasoning tasks. The gap is particularly pronounced in non-Latin languages. These findings suggest that existing benchmarks may overestimate LLMs’ true capacity to process and reason over extremely long input sequences \cite{michelangelo}. 

Apart from these design limitations, existing multilingual benchmarks also suffer from certain practical constraints that hinder their utility for long-context evaluation. For instance, both MLNeedle \cite{hengle2024_multilingualneedlehaystackinvestigating}, and mLongRR \cite{mLongRR} are prone to risks of data leakage during pretraining since they are curated using texts from open-source data. This is important as \textit{evaluation leakage} \footnote{Evaluation leakage occurs when the model has either fully or partially seen the evaluation (test) set during pretraining, which may happen because of poor data splits or data being publicly available. Evaluation leakage may result in misleadingly high-performance scores.} or \textit{short-circuiting} \footnote{Short-circuiting happens when a model doesn't actually infer the target task (in this case, reasoning over long contexts); but rather "cheats" using parametric knowledge or other shortcuts. Short-circuiting can inflate and compromise the integrity of results, making evaluation less reliable \cite{lee2024longcontextlanguagemodelssubsume}.}. 

\paragraph{\textbf{Contributions}}
To address these challenges, we introduce Multilingual Long-Context Reasoning (\dataset), a synthetic benchmark for evaluating LLM's long-context understanding beyond retrieval and haystacks. \dataset\ spans seven languages and integrates different tasks like retrieval, multi-hop inference, aggregation, and epistemic reasoning over extended multilingual inputs. \dataset\ can be scaled to any arbitrary context length, prevents evaluation leakage, and offers a more complete test of LLM's ability to reason over long multilingual inputs. A key methodological advantage of \dataset\ is that it is resistant to evaluation leakage and short-circuiting.  The instances in \dataset\ are generated synthetically using controlled and reproducible procedures, allowing the dataset to be extended to arbitrary context lengths without relying on pre-existing or web-sourced content. This ensures that the evaluation remains realistic and grounded in natural language while eliminating the risk of overlap with pretraining data. Furthermore, rather than focusing solely on surface-level retrieval, tasks in \dataset\ require reasoning over multiple facts, implicit relationships, multi-hop connections, temporal or spatial awareness, and so on. These task categories are \textit{minimal yet orthogonal}\footnote{As defined by \citet{michelangelo}, “minimal yet orthogonal” evaluation is a design approach where only a small set of tasks, each targeting a different and complementary aspect, can be used for a holistic evaluation. Here, the goal is to avoid redundancy or unnecessary complexity while still capturing a broad range of model performance.}. Thus, \dataset\ offers a complete but non-redundant view of an LLM's reasoning capabilities. 

In summary, this study makes the following contributions:
\begin{itemize}[itemsep=0pt]
    \item We propose \dataset, a synthetic benchmark to evaluate the multilingual long-context behaviour of LLMs. Going beyond surface-level retrieval, \dataset\ includes complex reasoning tasks in seven languages. For each language, \dataset\ provides $1,000$ QA instances evaluated across eight context lengths — from baseline (no distractors) to $128k$ tokens, a total of $8,000$ evaluation prompts per language.
    \item Through extensive experiments, we study how changes in the language, task, and prompting method affect the LLM's ability to reason over long multilingual contexts. Our results reveal a significant performance gap between high-resource and low-resource languages, particularly for tasks involving multi-step reasoning or resolving uncertainty. 
    \item We undertake a comparative analysis of retrieval and reasoning tasks, finding that existing benchmarks may overstate LLMs’ true long-context reasoning abilities. Empirically, we also observe that current models effectively utilize less than 30\% of their advertised context window for reasoning tasks.  
    \item To further support our findings, we conduct a detailed ablation study examining the impact of various hyperparameters, such as type of noise and sampling strategy, on the LLM's ability to reason over long contexts.
    \item We open-source the evaluation pipeline, source code, and the full  \dataset\ dataset to support future research on long-context evaluation and training of multilingual LLMs.
\end{itemize}
\section{Dataset}

\begin{figure}[t]
\centering
\includegraphics[width=\textwidth,keepaspectratio,trim={0 0 0 0},clip]{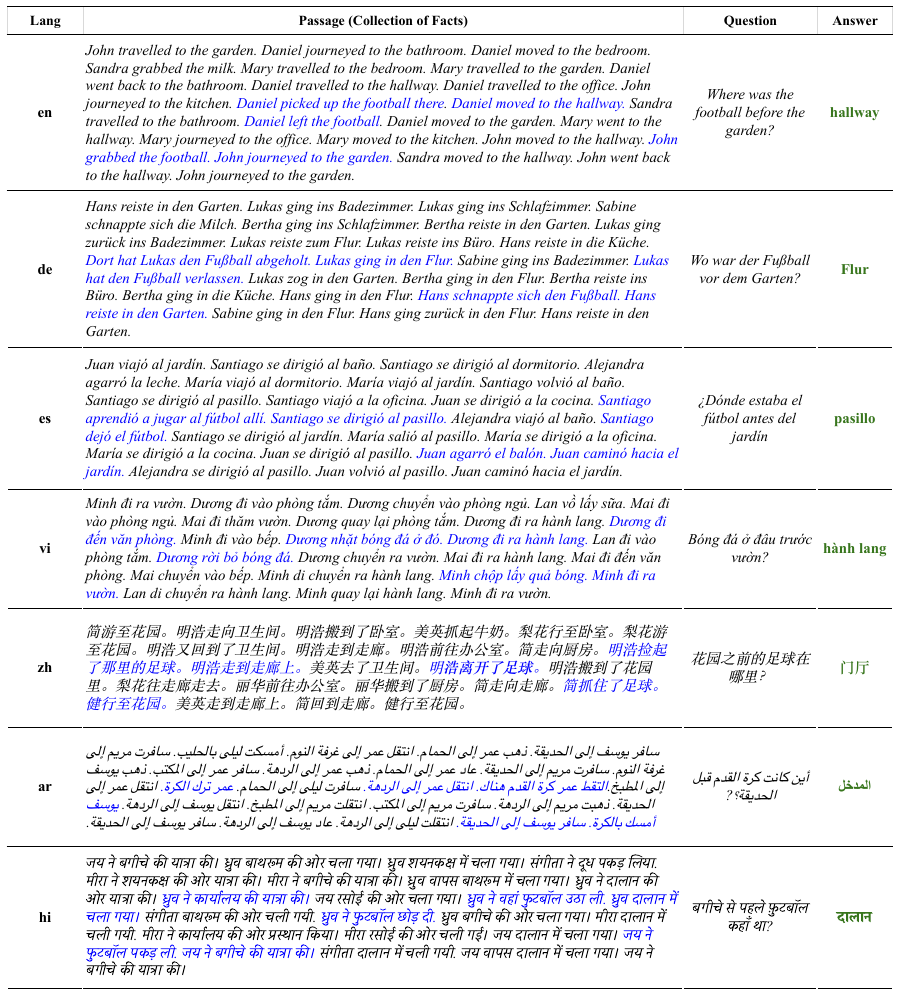}
\caption{An example from our proposed \dataset\ dataset. The original passage, question, and target from bAbI \cite{weston2015_babi} are translated from English to six other languages -- Spanish, German, Chinese, Vietnamese, Arabic, and Hindi. Thus, \dataset\ supports a parallel data structure where each dataset instance is available across all languages. This makes it ideal for equitable multilingual evaluation. }
\label{fig:dataset_example}
\vspace{-8mm}
\end{figure}

We curate \dataset\ as a multilingual, long-context adaptation of the bAbI dataset \cite{weston2015_babi}. bAbI provides a suite of English QA tasks to evaluate core reasoning abilities such as temporal and spatial awareness, association, counting, induction, and so on. As shown in Figure \ref{fig:dataset_example}, each bAbI data point consists of a passage, a question, and an answer. The passage is a sequence of independent statements (facts) -- each simulating an interaction between a ``character'' and an ``object'' such as ``\textit{John took the football to the garden}'' or ``\textit{Mary went to the kitchen}''. The accompanying questions are designed to test different aspects of reasoning, e.g., spatial awareness (``\textit{Where was the football before the garden?}'') or counting (``\textit{How many people visited the garden?}''). Therefore, the bAbI setup requires models to understand the interplay of different facts and to draw inferences based on the overall context as it unfolds throughout the passage. In this section, we detail the data curation process and tasks in \dataset.

\subsection{Design Principles}
\label{sec:design_principles}

We construct \dataset\ based on five design principles as outlined below.

\begin{itemize}[itemsep=5pt, label={}, leftmargin=0pt]
\item 1) \textbf{Parallelism}: \dataset\ is designed to be highly parallel -- each task instance is available across all the selected languages with aligned QA pairs. Such a parallel data structure enables reliable cross-lingual comparisons as it decouples language-specific performance from sample-level difficulty \cite{lewis2020-MLQA}. This makes the evaluation results more interpretable and consistent across languages.

\item 2) \textbf{Leakage and short-circuiting:} As discussed in Section \ref{sec:introduction}, evaluation benchmarks often face risks of evaluation leakage and short-circuiting. To address this, we construct \dataset\ using synthetically generated texts, which substantially lowers any chance of data leakage during pretraining. Additionally, all tasks in \dataset\ have non-trivial difficulty, which prevents LLMs from relying on superficial patterns or memorized knowledge to answer correctly, thereby reducing the risk of short-circuiting.

\item 3) \textbf{Minimal yet orthogonal tasks:} \dataset\ is designed with a minimal yet orthogonal approach \cite{michelangelo}. It includes a set of core tasks, each targeting a distinct aspect of long-context reasoning, such as retrieval, multi-hop inference, aggregation, and uncertainty handling. This design has two key advantages: (i) it enables broad evaluation of long-context understanding without overlap, and (ii) it ensures modularity so that future versions of the dataset can easily incorporate new reasoning tasks.

\item 4) \textbf{Domain-relevant distractors:} If background texts (i.e., distractors) are entirely out-of-distribution, they can make the evaluation \textit{a priori identifiable} \footnote{In long-context evaluation, \textit{a priori identifiable} means that the relevant information (or facts) can be identified purely from surface-level cues, without requiring any deeper understanding of how that information relates to the final task \cite{michelangelo}.}, and the task artificially easy. To avoid this, distractors in \dataset\ are designed to be in-distribution, i.e., they closely resemble the structure and content of the relevant information rather than being obviously unrelated. This helps prevent the tasks from turning into a simple pattern-matching problem, where the correct passages or facts stand out too clearly. \dataset\ offers fine-grained control over the type of distractors used, supporting three modes: (i) synthetically generated, (ii) sampled from external text corpora, and (iii) random noise.

\item 5) \textbf{Scalability:} \dataset\ is scalable to any arbitrary context length by simply varying the number and placement of distractor passages. Thus, \dataset\ can be used to test any current or new open-weight of API-based LLMs without changing the underlying task complexity. 

\end{itemize}

\subsection{Languages}
\label{sec:languages}

\dataset\ extends to seven typologically diverse languages: English (en), German (de), Spanish (es), Hindi (hi), Arabic (ar), Vietnamese (vi), and Simplified Chinese (zh). These languages vary substantially in terms of resource availability, ranging from high- to low-resource settings, as outlined by \citet{joshi-etal-2020-state}. They span four major language families: \textit{Indo-European} (en, es, de, hi), \textit{Afro-Asiatic} (ar), \textit{Austro-Asiatic} (vi), and \textit{Sino-Tibetan} (zh), and multiple writing systems — including Latin, Arabic, Devanagari, and logographic scripts. Our language selection also aligns with widely adopted multilingual evaluation benchmarks such as MLQA \cite{lewis2020-MLQA}, mMARCO \cite{bonifacio2022-mMARCO}, and MLNeedle \cite{hengle2024_multilingualneedlehaystackinvestigating}. Thus, our proposed dataset allows for (i) a systematic analysis of how script variation and linguistic typology affect long-context comprehension in LLMs, (ii) examining the impact of training data availability across languages with differing resource levels, and (iii) a direct comparison to existing multilingual benchmarks.

\subsection{Task Categories}
\label{sec:task_categories}

\dataset\ includes multiple reasoning tasks from bAbI covering four task categories -- retrieval, multi-hop inference, aggregation, and handling uncertainty. Below, we outline each category and the corresponding tasks. Detailed examples of each task are provided in Appendix \ref{tab:task-types}. 

\begin{itemize}[itemsep=3pt, leftmargin=10pt]

\item \textbf{Retrieval.} This category follows the needle-in-a-haystack framework \cite{gregory2023-NIAH}, and broadly tests the LLM's \textit{associative recall} over the input context. Here, we include two tasks from the original bAbI dataset: 

\begin{itemize}[noitemsep,topsep=1pt, after=\vspace{0pt}]
  \item \textbf{Basic factoid QA:} Task 1 consists of a set of questions where the answer is present as a single supporting fact among two or more irrelevant facts. The model must locate one key sentence among possibly many. The questions are framed in the \textit{WhereIsActor} and \textit{WhoWhatGave} formats \cite{weston2015_babi}, respectively. 
  \item \textbf{Yes/No or negation:} Task 2 consists of a set of questions where the answer is binary (yes/no), and can be directly inferred from a single supporting fact within the passage. The questions are framed in the format \textit{IsActorThere} \cite{weston2015_babi}, and test the model's ability to identify true/false statements or simple negation. 
\end{itemize}

\item \textbf{Multi-hop inference:} Multi-hop inference requires the LLM to combine two or more pieces of information (facts) from the context in order to arrive at the answer. This includes temporal, logical, and spatial chaining, where each fact contributes to the final inference. 

\begin{itemize}[noitemsep,topsep=0pt, after=\vspace{0pt}]
  \item \textbf{Fact chaining:} Task 3 consists of questions where the answer depends on two or more facts chained across multiple sentences, often requiring tracking over time. The model is required to track a chain of characters and their actions sequentially, for instance, where an object moved and what happened before a specific event.
  \item \textbf{Argument relations:} Task 4 consists of questions where the answer has to be inferred by understanding the relationship or interplay between characters. The model must reason over relationships (e.g., spatial directions or entity roles) rather than follow character movements.
\end{itemize}

\item \textbf{Aggregation.}
Aggregation tasks test the language models' ability to compile or summarise different pieces of information. This includes counting specific entities or compiling sets of items across the context. Unlike multi-hop reasoning, aggregation focuses on synthesis rather than chaining \cite{michelangelo}.

\begin{itemize}[noitemsep,topsep=1pt, after=\vspace{0pt}]
  \item \textbf{Counting:} In Task 5, the questions require counting the number of relevant entities or objects based on events in the passage. The model is required to track state changes across multiple characters and actions, making it challenging.
  \item \textbf{List / Sets:} In Task 6, the answer is a set or list of entities retrieved from the context, often based on participation or possession. The language model must identify and maintain character and object states (e.g., what someone picked up and didn’t drop).
\end{itemize}

\item \textbf{Uncertainty.}
Uncertainty or epistemic reasoning evaluates the language models' ability to recognize ambiguity or incomplete knowledge and respond appropriately. Rather than guessing, the model must either eliminate impossibilities or refrain from over-committing to ambiguous inputs.

\begin{itemize}[noitemsep,topsep=1pt, after=\vspace{0pt}]
  \item \textbf{Indefinite knowledge:} In Task 7, the question is about a fact that the story makes ambiguous or uncertain. The model must recognize uncertainty, eliminate false options, and avoid over-committing to ambiguous information.
\end{itemize}

\end{itemize}

\subsection{Translation Process}


\begin{figure}[t]
\centering \includegraphics[width=0.7\columnwidth]{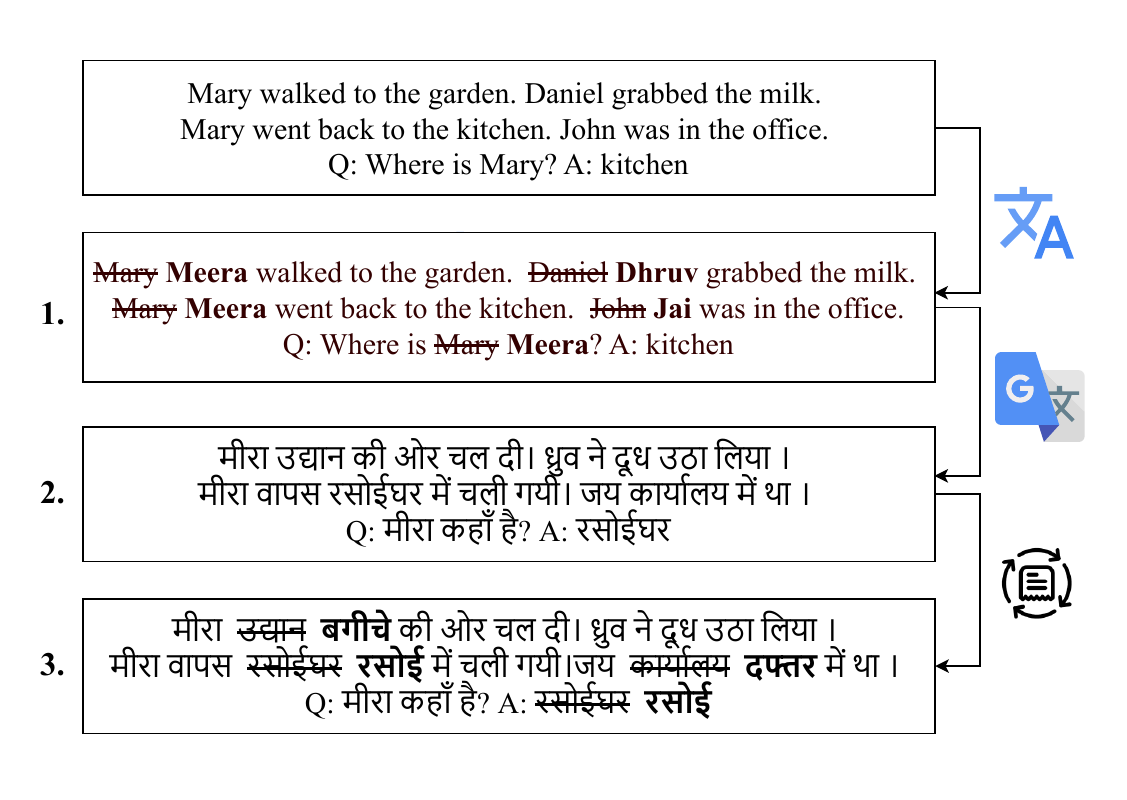}
\caption{The three-step process used to translate the \textit{passage}, \textit{question}, and \textit{answer} from bAbI to its Hindi equivalent form. The same process is followed for other selected languages from Table \ref{fig:dataset_example}. }
\label{fig:translation_process}
\vspace{-5mm}
\end{figure}

We begin with the extraction of independent facts from the original English bAbI dataset. From an initial pool of $1k$ examples, approximately $8,500$ unique facts are identified. The passage, question, and answer are then translated into six languages, as described in Section \ref{sec:languages}, resulting in a parallel multilingual dataset. As illustrated in Table \ref{fig:dataset_example}, this structure supports direct cross-lingual comparison and aligns with the format of existing multilingual benchmarks such as MLQA \cite{lewis2020-MLQA} and MLNeedle \cite{hengle2024_multilingualneedlehaystackinvestigating}. To make the dataset more representative across languages, we replace character names in bAbI with culturally appropriate alternatives for each target language, as shown in Table \ref{tab:crosslingual_names}. Following this pre-processing step, we use the Google Translate API\footnote{https://cloud.google.com/translate/docs/reference/api-overview - Google Translate API} to translate the passage, question, and answer from English to all selected languages \footnote{Note that we translate independent sentences and not the entire text at once, as detailed in Appendix \ref{appendix:translation}}. Figure \ref{fig:translation_process} provides an example of the three-step translation process. The resulting dataset preserves structural consistency and ensures that the semantics of each passage, question, and answer remain intact across all seven languages. We defend the absence of human translation in Appendix \ref{tab:translation_QE}. 

\subsection{Background Texts}
\label{sec:distractors}

Background texts, or \textit{distractors}, refer to segments of long-context that are irrelevant to the target task or set of facts the model is expected to reason over. Following prior work by \citet{gregory2023-NIAH} and \citet{kuratov2024babilongtestinglimitsllms}, our long-context prompts are constructed by interleaving relevant facts with multiple distractor passages. This setup mimics real-world cases where relevant information is often embedded within unrelated content. To preserve the effectiveness of distractors, it is important that they are not completely out-of-distribution with the task, as this can risk making the relevant information overly salient and undermine the purpose of evaluating a model’s ability to filter from noise \cite{michelangelo}. To mitigate this, we employ semantic similarity as a filtering mechanism for distractor selection. Specifically, given a task instance (passage, query) and a corpus of background texts, we retrieve the $top$-$k$ most semantically similar distractor passages based on cosine similarity. The value of $k$ is adjusted dynamically to control the overall length of the prompt, allowing our setup to scale the long contexts to any arbitrary length as needed. As a source of background texts, we experiment with three types of distractors: (i) \textit{synthetic distractors} generated using GPT-4, (ii) \textit{natural distractors} sampled from a public text corpus, and (iii) \textit{random noise} to serve as a control condition. Below, we describe each of them briefly. 

\paragraph{\textbf{Synthetic distractors}}
Recent work, such as \citet{hsieh2024-Ruler}, has explored the use of synthetically generated distractors to control task complexity and make evaluation \textit{a posteriori identifiable} \footnote{If relevant information is \textit{a posteriori identifiable}, it means that the model has to understand and reason over the information before arriving at an answer.}. Building on this idea, we construct our own corpus of in-domain synthetic distractor texts using GPT-4 \cite{gpt4-technical-report}. As illustrated in Table \ref{fig:dataset_example}, passages in \dataset\ are a collection of independent factual statements. In total, \dataset\ includes $8,500$ unique facts, each following a common grammatical structure: subject (character) → action (verb) → object (or location/entity). These facts serve as seed examples for generating distractors. For each fact, we prompt GPT-4 to generate a short passage or story that is structurally similar (in terms of style, grammar, or tone) but semantically irrelevant to both the original passage and query. To introduce variety, we use four different prompt templates along with three levels of temperature sampling ($0$, $0.5$, and $1.0$). The specific templates used in this process are shown in Figures \ref{fig:distractor_prompt_1} - \ref{fig:distractor_prompt_4}.

\paragraph{\textbf{Natural distractors}} Many existing benchmarks source distractor texts from publicly available corpora such as Wikipedia articles, Paul Graham essays, or books. In this study, we use natural long-form passages from the MLQA dataset \cite{lewis2020-MLQA} as our source of distractors. MLQA is well-suited for our multilingual setting, as it provides parallel passages across all seven languages included in \dataset.

\paragraph{\textbf{Random noise}} The distractor texts are random character-level tokens that carry no semantic meaning, grammatical structure, or lexical relevance to the primary task. Their purpose is purely to inflate context length without introducing any interpretive content. We use this as a control condition for long-context evaluation.
\begin{figure}
    \centering
    \includegraphics[width=\linewidth]{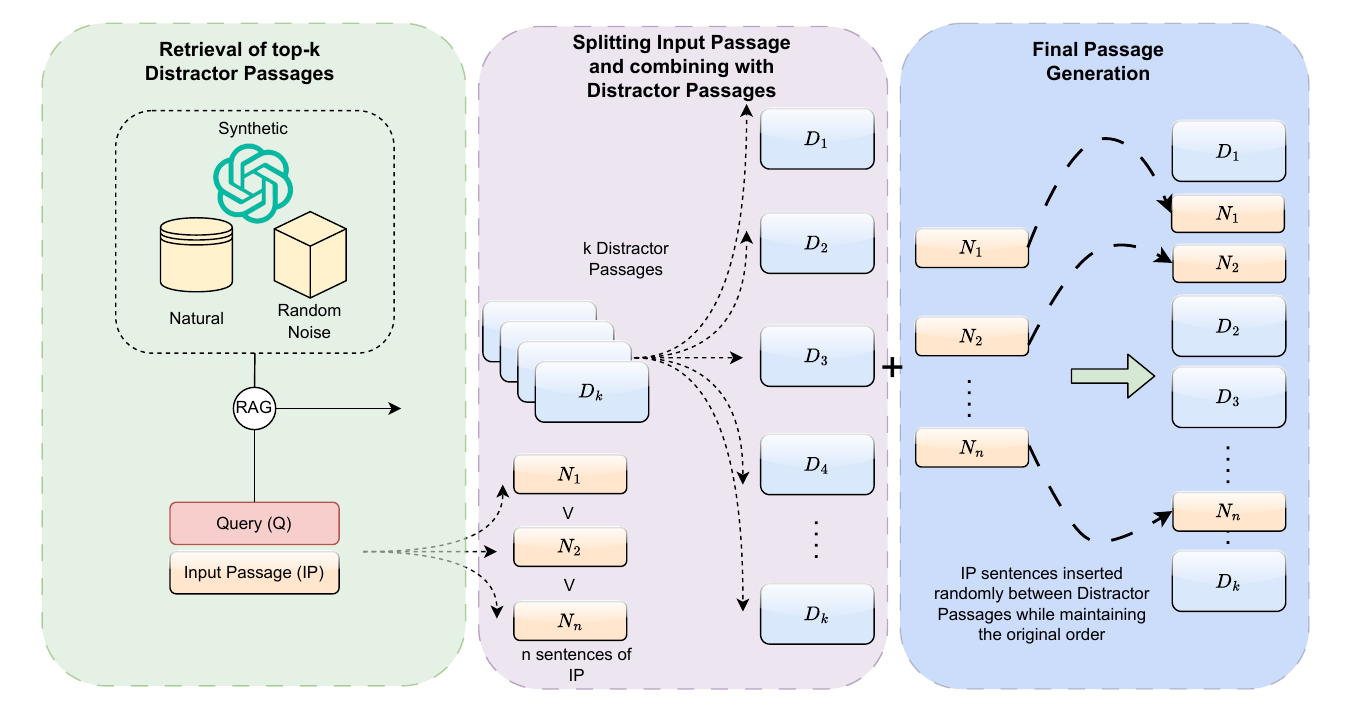}
    \caption{Overview of our pipeline for creating \texttt{MLRBench}. An Input Passage ($IP$) is split into $n$ sentences. Given $IP$ and the query, top-k distractor passages are retrieved from sources -- (1) synthetic passages from GPT-4, (2) public datasets and (3) random token-level noise. The independent sentences from $IP$ are randomly inserted at positions between distractor passages but maintain their original order. The value $k$ is dynamically controlled to increase or decrease the context size of the final passage.}
    \label{fig:final_passage_generation}
    \vspace{-6mm}
\end{figure}
\section{Experimental Setup}

\subsection{Task Overview}
\label{sec:task_overview}

In this section, we describe the construction of our long-context prompt (\( P \)) for evaluating our task. Our structure is similar to the widely used {needle-in-a-haystack} paradigm \cite{Liu2023-LostITM}, where sentences from a relevant passage \textbf{(\( IP \))}  are dispersed within the set of distractor passages (\( D \)) while preserving their original chronological order. As shown in Figure \ref{fig:dataset_example}, each data point in our experiment is derived from the original bAbI dataset and consists of:



\begin{itemize}[itemsep=0pt, leftmargin=10pt]
    \item {Input Passage \textbf{(\( IP \))} }: A structured text composed of \( n \) sequentially-ordered independent sentences.
    \item {Question (\( Q \))}: A query related to the information within \( IP \).
    \item {Ground-truth Answer (\( A \))}: The correct response to \( Q \).
\end{itemize}


\noindent
We formally define the input passage as:
\begin{equation}
    IP = \{N_i \mid i = 1, 2, \dots, n\},
\end{equation}
where each \( N_i \) represents an independent sentence within \( IP \), following the strict chronological ordering constraint:
\begin{equation}
    N_1 \prec N_2 \prec \dots \prec N_n,
\end{equation}
where \( \prec \) denotes that sentence \( N_i \) must appear before \( N_{i+1} \) in the final prompt (see Figure \ref{fig:dataset_example}). Additionally, we define the set of \( m \) unique distractor passages as:
\begin{equation}
    D = \{D_j \mid j = 1, 2, \dots, m\},
\end{equation}
where each \( D_j \) is an independent distractor passage from a synthetic corpus generated using an LLM.

\paragraph{\textbf{Constructing long-context}}

The long context \( P \) is constructed by placing sentences in \( IP \) randomly within the distractor passages \( D \). This is done while maintaining the original order of sentences in \( IP \). Formally, the sentences \( \{N_1, N_2, \dots, N_n\} \) from \( IP \) must retain their original ordering within \( P \).
Let $\pi$ be an injective map from the set $\{1, 2,\ldots, n\}$ to the set $\{1, 2, \ldots, m\}$ such that
\begin{equation}\label{eq:ordering_constraint}
    \pi(i) < \pi(j) \hspace{1em}\forall\hspace{1em} \{i < j : 1 \leq i, j \leq n\}
\end{equation}
where \( \pi(i) \) represents the position of sentence \( N_i \) in \( P \), and \( D \) provides the background text, acting as the "haystack". Thus, our setup allows one to dynamically increase the "haystack" to any arbitrary length by adjusting the number of distractor passages \( m \). 

\paragraph{\textbf{Final prompt}}

The final prompt $P = \{P_1,P_2, ..., P_{m+n}\}$ is constructed as follows:
\begin{equation}
P_k = 
\left\{
    \begin{array}{lr}
        N_i, & \exists i : \pi(i) = k\\
        D_j \in D, & \text{otherwise}
    \end{array}
\right\}
\end{equation}
where:
\begin{itemize}[itemsep=1pt]
    \item \( D_{j} \) is sampled from \(D\) without replacement.
    \item $IP$ = \(\{ N_1, N_2, \dots, N_n \}\) remain interspersed throughout \( D\) but retain their relative order due to the property in Equation \ref{eq:ordering_constraint}.
    \item The total length of \( P \) is significantly larger than \( IP \) alone, increasing the complexity of the retrieval.
\end{itemize}

\subsection{Language Models}
We conduct all our experiments using the open-weight \text{Llama 3.1-Instruct}\footnote{We employ the \textbf{meta-llama/Llama-3.1-8B-Instruct} model checkpoints from huggingface.} model, which is an instruction-finetuned version of Llama 3.1 \cite{Llama3.1}. Llama3.1-Instruct supports a context-size of $128k$ tokens and is shown to perform exceedingly well on multiple multilingual tasks, owing to its use of optimised grouped query attention \cite{chen2024_optimisedgroupedqueryattentionmechanism}. For our RAG experiments, we make use of an off-the-shelf and a specialised multilingual models -- Jina-reranker (\text{JinaAI}) \cite{jinarerankerv2} and paraphrase-multilingual-mpnet-base-v2 (\text{MPNet}) \cite{reimers-2020-multilingual-sentence-bert}, respectively. 

\subsection{Baselines}
\begin{table}[!]
\begin{center}
\resizebox{\textwidth}{!}{%
\begin{tabular}{c|l|cccccccc|c|c|c}
\toprule
\textbf{Method} & \textbf{Prompt Type} & \textbf{0\textit{k}} & \textbf{2\textit{k}} & \textbf{4\textit{k}} & \textbf{8\textit{k}} & \textbf{16\textit{k}} & \textbf{32\textit{k}} & \textbf{64\textit{k}} & \textbf{128\textit{k}} & \textbf{Mean LC} & \textbf{Drop (\%)} & \textbf{ECW } \\
\midrule
\multirow{4}{*}{\shortstack{Prompting\\Baselines}} & Zeroshot & $0.513$ & $0.406$ & $0.405$ & $0.366$ & $0.357$ & $0.325$ & $0.311$ & $0.317$ & \cellcolor{white!20}$0.318$ & \cellcolor{white!50}$8.89$ & $4K$ \\
& Translation & $0.540$ & $0.414$ & $0.403$ & $0.384$ & $0.368$ & $0.331$ & $0.316$ & $0.323$ & \cellcolor{white!30}$0.324$ & \cellcolor{white!50}$9.03$ & $4K$ \\
& CoT & \underline{\textbf{$0.598$}} & $0.481$ & $0.477$ & $0.439$ & $0.448$ & $0.382$ & $0.285$ & $0.253$ & \cellcolor{white!10}$0.307$ & \cellcolor{white!75}$22.77$ & $16K$ \\
& Fewshot & $0.521$ & $0.468$ & $0.433$ & $0.408$ & $0.423$ & $0.348$ & $0.333$ & $0.334$ & \cellcolor{white!40}$0.338$ & \cellcolor{white!70}$13.43$ & $16K$ \\
\midrule
\multirow{4}{*}{\shortstack{RAG\\(JinaAI, \textit{top-k}=100)}} & Zero-shot & - & $0.429$ & $0.397$ & $0.381$ & $0.340$ & $0.334$ & $0.330$ & $0.300$ & \cellcolor{white!35}$0.321$ & \cellcolor{white!60}$12.91$ & $8K$ \\
& Translation & - & $0.435$ & $0.390$ & $0.388$ & $0.364$ & $0.368$ & $0.349$ & $0.342$ & \cellcolor{white!50}$0.353$ & \cellcolor{white!40}$9.31$ & $8K$ \\
& CoT & - & $0.473$ & $0.455$ & $0.443$ & \underline{\textbf{$0.446$}} & $0.438$ & $0.412$ & $0.400$ & \cellcolor{white!60}$0.417$ & \cellcolor{white!20}$7.26$ & $32K$ \\
& Few-shot & - & $0.473$ & $0.455$ & $0.443$ & $0.440$ & $0.431$ & $0.403$ & $0.393$ & \cellcolor{white!65}$0.409$ & \cellcolor{white!30}$7.97$ & $32K$ \\
\midrule
\multirow{4}{*}{\shortstack{RAG\\(MpNet, \textit{top-k}=100)}} & Zeroshot & - & $0.424$ & $0.405$ & $0.388$ & $0.378$ & $0.366$ & $0.339$ & $0.337$ & \cellcolor{white!55}$0.347$ & \cellcolor{white!50}$8.74$ & $8K$ \\
& Translation & - & $0.435$ & $0.412$ & $0.387$ & $0.401$ & $0.393$ & $0.357$ & $0.349$ & \cellcolor{white!70}$0.366$ & \cellcolor{white!40}$8.57$ & $8K$ \\
& CoT & - & \underline{\textbf{$0.473$}} & \underline{\textbf{$0.462$}} & \underline{\textbf{$0.449$}} & $0.445$ & \underline{\textbf{$0.439$}} & \underline{\textbf{$0.416$}} & \underline{\textbf{$0.414$}} & \cellcolor{white!70}\underline{\textbf{$0.423$}} & \cellcolor{white!10}$5.83$ & $32K$ \\
& Fewshot & - & $0.473$ & $0.462$ & $0.449$ & \underline{\textbf{$0.449$}} & $0.432$ & $0.409$ & $0.401$ & \cellcolor{white!75}$0.414$ & \cellcolor{white!20}$7.14$ & $32K$ \\
\bottomrule
\end{tabular}%
}
\end{center}
\caption{We highlight the overall performance (averaged across all languages) of the selected model. "Mean LC" denotes the mean long-context performance (averaged across context windows $32k$, $64k$, and $128k$), respectively. "Drop \%" denotes the percentage drop in accuracy when moving from the baseline (no token) to $100k$ tokens. Effective context window (ECW) is the context size up to which accuracy remains within 30\% of the baseline performance. While the model claims to support a long context size of $128k$ tokens, the effective context size is $32k$ at best.}
\label{tab:main_results}
\vspace{-6mm}
\end{table}
\label{sec:baselines}

\paragraph{\textbf{Prompting}}
We experiment with four prompting stategies -- \textit{zeroshot} (ZS), \textit{fewshot} (FS), \textit{chain-of-thoughts} (CoT), and \textit{in-context translation} (ICT). ICT is a type of cross-lingual prompting in which the model is instructed to interpret the context and respond in English. Translating non-English prompts into English before inference has been shown to improve performance on certain multilingual tasks \cite{ICT3, ICT1}. ICT is similar to the widely adopted practice of \textit{pre-translation} \cite{ICT0}, except that translation happens in-context, and thus, it avoids any additional overhead or information loss \cite{ICT2, ICT4}. 

For all our experiments, we follow a standard Instruction $\rightarrow$ Context $\rightarrow$ Question prompt format. Each evaluation prompt begins with a task-specific instruction, as defined in Table \ref{tab:task-instructions}. In-context demonstrations for few-shot prompting are retrieved via semantic matching (\textit{top-k} selection) using sentence-transformers\footnote{We use the multilingual Sentence-BERT model to retrieve \textit{top-k} instances from the dev set, such that they are most relevant to both the passage and question while ensuring they do not contain the answer.} \cite{reimers-2020-multilingual-sentence-bert}. As context and few-shot examples form the main body of the prompt, they are enclosed under the <context></context> and <example></example> tags, respectively. To avoid any post-processing step, we force the LLM to format its output in JSON. Prompt templates used for ZS, ICT, CoT, and FS are defined in Figures \ref{fig:zs_prompt_template}, \ref{fig:cot_prompt_template}, and \ref{fig:fs_prompt_template} respectively. 

\paragraph{\textbf{Retrieval Augmented Generation (RAG)}}

RAG provides a way to overcome the fixed-context windows of LLMs by dynamically retrieving and re-ranking relevant content from a larger context \cite{rag-LLMs}. In this work, we experiment with two off-the-shelf RAG retriever models: Jina-reranker (\text{JinaAI}) \cite{jinarerankerv2} and multilingual-mpnet-base-v2 (\text{MPNet}) \cite{reimers-2020-multilingual-sentence-bert}. We adopt a simple RAG pipeline based on $top-k$ retrieval. As defined in Section \ref{sec:task_overview}, an evaluation prompt $P$ consists of $n$ input passages$IP$ interspersed within $m$ distractor texts $D$. The retriever models encode and re-rank the sentences within $P$ based on their relevance to the question $Q$. Subsequently, a new condensed prompt is constructed using the $top-k$ most relevant sentences with respect to $Q$, while preserving the original chronological order of sentences from both $IP$ and $D$. Finally, this new RAG prompt is used for inference experiments.

\subsection{Evaluation Metrics}
\label{sec:evaluation_metrics}
We use \textbf{exact accuracy} as our primary evaluation metric. Following prior work \cite{wang-Multimodal-NIAH-2024, hengle2024_multilingualneedlehaystackinvestigating}, exact accuracy is defined as the proportion of samples where the model’s predicted output includes the ground-truth answer in any of the target languages in \dataset. In essence, it checks whether the correct answer is present in the model’s prediction. Additionally, we evaluate RAG performance using \textbf{Recall@k}, which measures the \textit{proportion of relevant sentences retrieved for a given question}, calculated as the size of the intersection between the retrieved and relevant sets, divided by the total number of relevant sentences \cite{recall_metric}.
\begin{figure*}[t]
\includegraphics[width=\textwidth,keepaspectratio,trim={0 0 0 0}, clip]{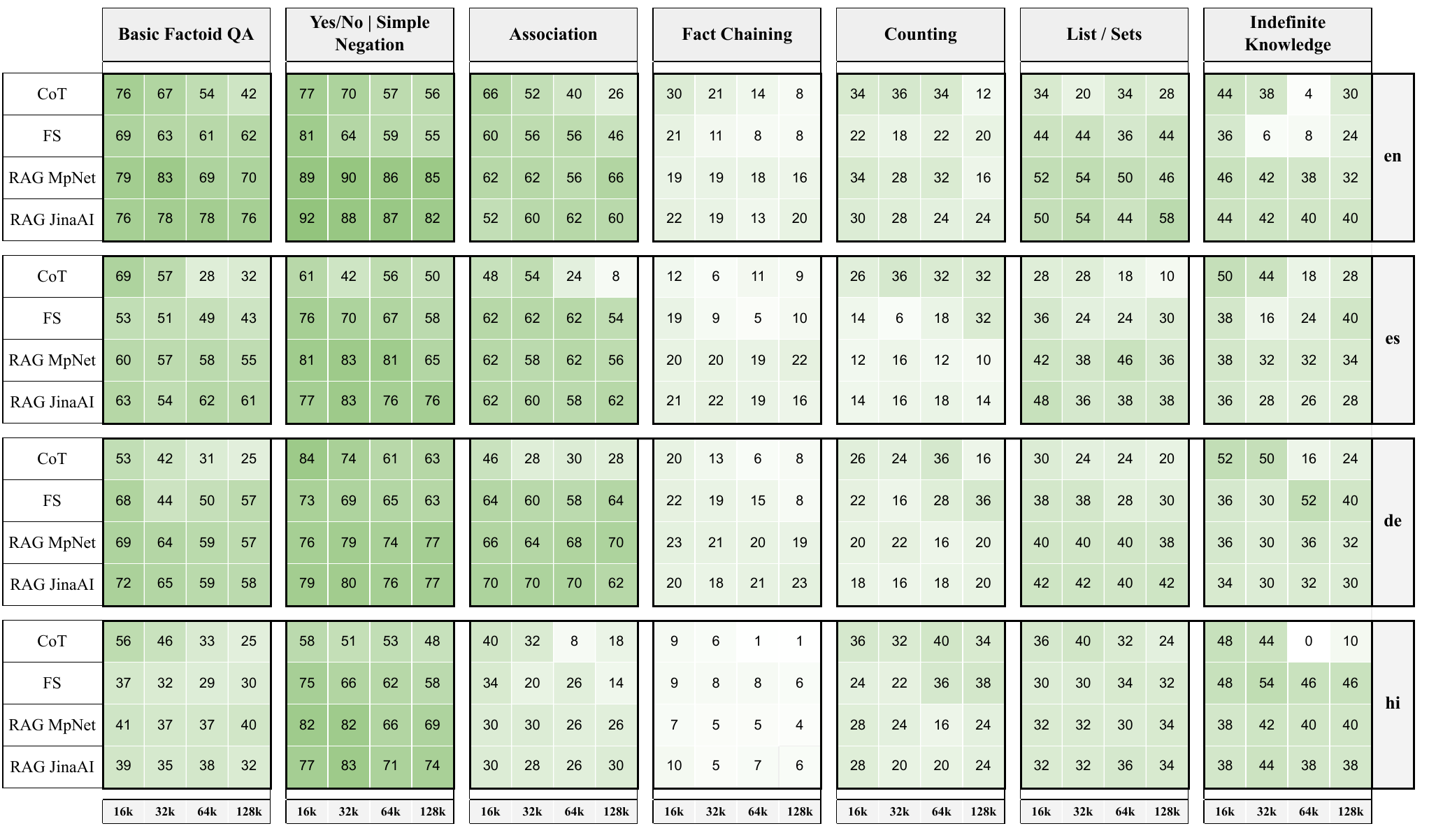}
\caption{Heatmap depicting performance for different tasks with increasing context window ($16k$ to $128k$). All values represent accuracy scores on a percentage scale, with higher values getting darker shades and vice versa.}
\label{fig:result_table5a}
\vspace{-5mm}
\end{figure*}

\begin{figure*}[t]
\includegraphics[width=\columnwidth,keepaspectratio,trim={0 0 0 0},clip]{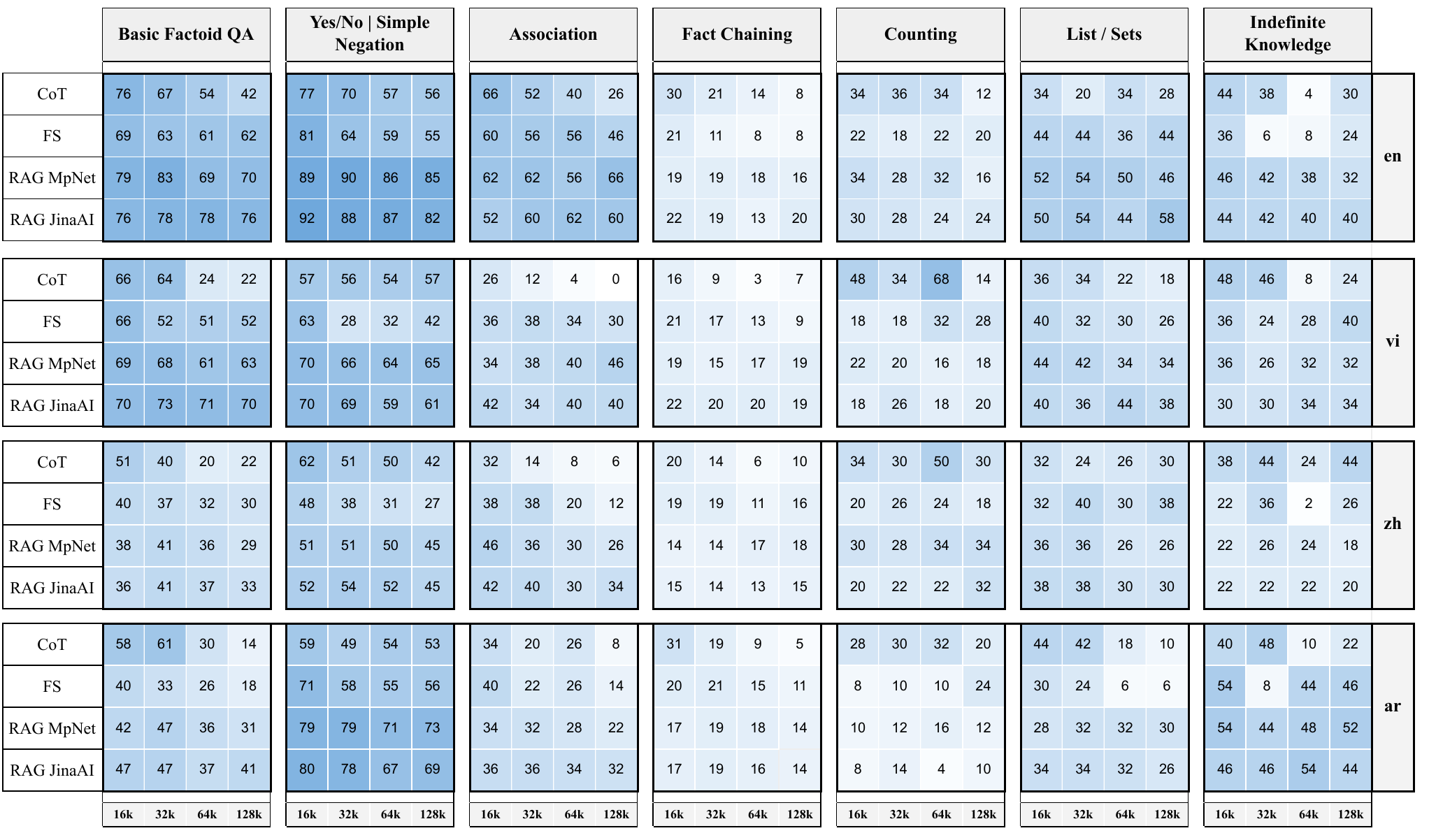}
\caption{Heatmap depicting performance for different tasks with increasing context window ($16k$ to $128k$). All values represent accuracy scores on a percentage scale, with higher values getting darker shades and vice versa.}
\label{fig:result_table5b}
 \vspace{-5mm}
\end{figure*}
\section{Experimental Results}
We conduct inference experiments across a range of context sizes, beginning with the baseline setup (no distractors) and extending up to $128k$ tokens. Table \ref{tab:main_results} presents a summary of results for all baselines, averaged across the seven languages. Notably, even the best-performing method at the baseline, i.e., CoT, achieves only a moderate $60\%$ success rate, suggesting that questions in \dataset\ have non-trivial difficulty even in the absence of distractors. Across all methods, we observe a consistent decline in accuracy as context length increases, with the degradation becoming especially pronounced at longer contexts beyond $32k$ tokens. In the following section, we present detailed results and highlight the key trends.

\paragraph{\textbf{Prompt-only methods struggle to maintain accuracy even at moderate contexts}}
Among prompt-only methods, CoT and FS, outperform ZS and ICT across all context lengths and languages consistently, with an accuracy margin of at least $6\%$ on average. However, prompt-based methods are also highly sensitive to an increase in context size. Their performance declines significantly, particularly beyond a context window of $16k$ tokens. For instance, CoT accuracy drops from $0.598$ at baseline to $0.253$ at $128k$, reflecting a relative $95\%$ performance loss. On average, prompt-based methods lose more than $50\%$ of their baseline accuracy at longer context windows ($\geq32k$), and nearly all prompting methods begin to deteriorate sharply after $8k$ to $16k$ tokens. This highlights the limited capacity of these methods to handle extended input contexts without additional retrieval mechanisms. 

\paragraph{\textbf{While RAG provides stability, it does not solve the long-context problem}}

Relative to their prompt-only counterparts, RAG-based methods show a smaller drop in accuracy between $2k$ and $128k$ tokens. For example, the relative decline for MpNet (CoT) is $40.41\%$, compared to $95.53\%$ decline for CoT alone. In addition, RAG-based methods show a flatter degradation slope, indicating improved robustness and more consistent performance as input context increases. We observe that without RAG, performance deteriorates rapidly after $4k–16k$ tokens for almost all prompting strategies. In contrast, RAG, when combined with CoT or FS prompting, achieves relatively stable performance as context length increases. Among the RAG methods, the best-performing model is MpNet combined with CoT, which attains a mean accuracy of $0.423$ across long contexts ($\geq32k$), the highest across all methods.

\begin{figure}
    \centering
    \includegraphics[width=0.8\linewidth]{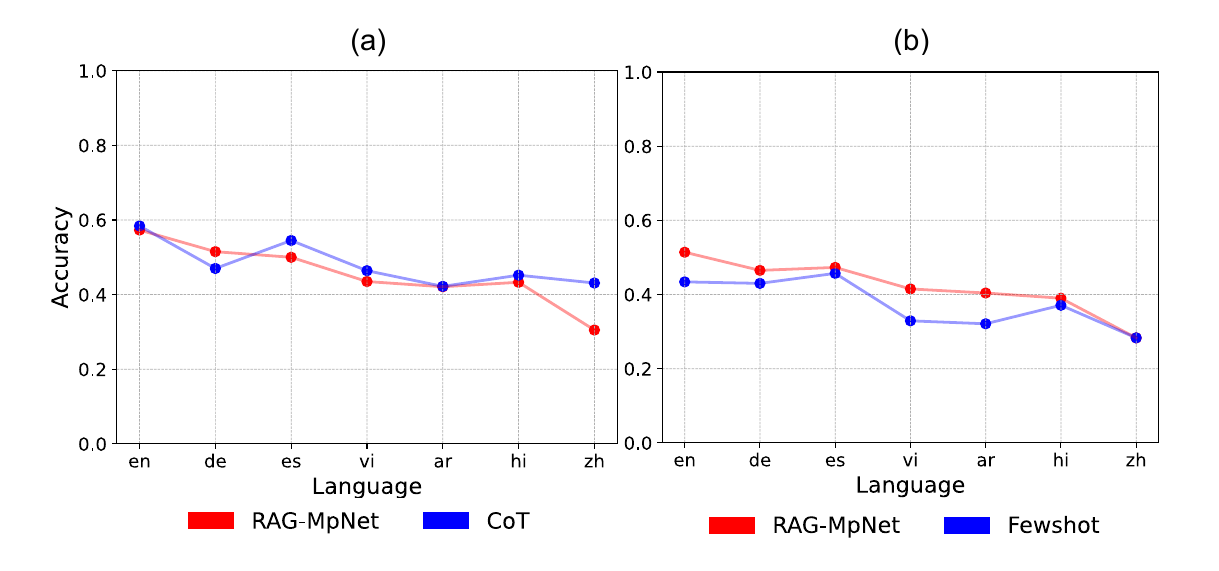}
    \caption{Performance comparison between the best RAG method and the best prompt-only method for (a) short context lengths ($4k-16k$) and (b) long context lengths ($\geq$ $32k$).}
    \label{fig:RS-RQ}
    \vspace{-6mm}
\end{figure}

Table \ref{tab: RS-RQ} shows the performance of different prompt-based and RAG-based methods for all languages across short context size and long context size, respectively. For both short and long context sizes, we observe that RAG-MpNet outperforms RAG-JinaAI. Among the prompt-based methods, CoT and FS perform best for short and long context sizes. We compare the performance of RAG-MpNet against the corresponding best prompt-based methods for all languages in Figure \ref{fig:RS-RQ}. For shorter context sizes, CoT predominantly outperforms RAG-MpNet, while for longer context sizes, RAG-MpNet outperforms FS prompting.

Despite maintaining performance with increased context lengths, RAG does not entirely mitigate the effects of long-context scaling. Even under optimal configurations, the accuracy of RAG-based methods at longer context windows remains approximately $20\%-40\%$ below the best short-context baseline. This indicates that while RAG improves stability and extends effective context size, there remains considerable room for further advancement in handling extreme context lengths.


\begin{table}[t]
\small
\begin{center}
\resizebox{\columnwidth}{!}{%
\begin{tabular}{l|l|ccccccc|c}
\toprule
\textbf{Language} & \textbf{\text{\textit{top-k}}} & \textbf{en} & \textbf{de} & \textbf{es} & \textbf{zh} & \textbf{vi} & \textbf{hi} & \textbf{ar} & \textbf{Avg.} \\
\midrule
\multirow{3}{*}{RAG-JinaAI} 
 & 20  & $0.423$ & $0.414$ & $0.385$ & $0.217$ & $0.376$ & $0.343$ & $0.289$ & $0.349$ \\
 & 100 & $0.399$ & $0.151$ & $0.357$ & $0.311$ & $0.389$ & $0.347$ & $0.328$ & $0.326$ \\
 & 300 & $0.405$ & $0.345$ & $0.350$ & $0.228$ & $0.313$ & $0.326$ & $0.310$ & $0.325$ \\
\midrule
\multirow{3}{*}{RAG-MpNet} 
 & 20  & $0.472$ & $0.414$ & $0.380$ & $0.204$ & $0.366$ & $0.308$ & $0.316$ & $0.351$ \\
 & 100 & $0.438$ & $0.405$ & $0.406$ & $0.225$ & $0.376$ & $0.320$ & $0.315$ & $0.355$ \\
 & 300 & $0.425$ & $0.362$ & $0.361$ & $0.223$ & $0.328$ & $0.315$ & $0.320$ & $0.333$ \\
\midrule
\multicolumn{2}{c|}{\textbf{MpNet - JinaAI (\textit{top-k}=100)}} & \textcolor{ForestGreen}{$\uparrow0.039$} & \textcolor{ForestGreen}{$\uparrow0.254$} & \textcolor{ForestGreen}{$\uparrow0.049$} & \textcolor{red}{$\downarrow0.086$} & \textcolor{red}{$\downarrow0.013$} & \textcolor{red}{$\downarrow0.027$} & \textcolor{red}{$\downarrow0.013$} & \textcolor{ForestGreen}{$\uparrow0.029$} \\
\bottomrule
\end{tabular}%
}
\end{center}
\caption{Performance of RAG-JinaAI and RAG-MpNet across different \textit{top-k} values for various languages. The last row represents the difference between RAG-MpNet and RAG-JinaAI at \textit{top-k}=$100$.}
\label{tab:rag_language_topk}
\vspace{-3mm}
\end{table}

\begin{table}[!t]
\small
\begin{center}
\resizebox{0.75\columnwidth}{!}{%
\begin{tabular}{lc|c|c|c|c|c|c}
\toprule
\textbf{Method} & \textbf{\textit{top-k}} & \textbf{4\textit{k}} & \textbf{8\textit{k}} & \textbf{16\textit{k}} & \textbf{32\textit{k}} & \textbf{64\textit{k}} & \textbf{128\textit{k}} \\
\midrule
\multirow{3}{*}{RAG-JinaAI} 
 & 20  & \cellcolor{white!20}$0.383$ & \cellcolor{white!20}$0.322$ & \cellcolor{white!20}$0.309$ & \cellcolor{white!30}$0.257$ & \cellcolor{white!40}$0.181$ & \cellcolor{white!40}$0.178$ \\
 & 100 & \cellcolor{white!60}$0.759$ & \cellcolor{white!60}$0.581$ & \cellcolor{white!60}$0.557$ & \cellcolor{white!50}$0.460$ & \cellcolor{white!50}$0.329$ & \cellcolor{white!50}$0.320$ \\
 & 300 & \cellcolor{white!90}$0.989$ & \cellcolor{white!90}$0.950$ & \cellcolor{white!90}$0.847$ & \cellcolor{white!80}$0.696$ & \cellcolor{white!70}$0.471$ & \cellcolor{white!70}$0.453$ \\
\midrule
\multicolumn{2}{c|}{$\Delta_{\mathrm{max - min}}$} & \textcolor{red}{$\downarrow0.606$} & \textcolor{red}{$\downarrow0.628$} & \textcolor{red}{$\downarrow0.537$} & \textcolor{red}{$\downarrow0.439$} & \textcolor{red}{$\downarrow0.290$} & \textcolor{red}{$\downarrow0.275$} \\
\midrule
\multirow{3}{*}{RAG-MpNet} 
 & 20  & \cellcolor{white!20}$0.517$ & \cellcolor{white!20}$0.431$ & \cellcolor{white!30}$0.412$ & \cellcolor{white!30}$0.345$ & \cellcolor{white!40}$0.248$ & \cellcolor{white!40}$0.243$ \\
 & 100 & \cellcolor{white!70}$0.926$ & \cellcolor{white!70}$0.809$ & \cellcolor{white!70}$0.776$ & \cellcolor{white!60}$0.635$ & \cellcolor{white!60}$0.446$ & \cellcolor{white!60}$0.435$ \\
 & 300 & \cellcolor{white!90}$0.989$ & \cellcolor{white!90}$0.996$ & \cellcolor{white!90}$0.980$ & \cellcolor{white!80}$0.905$ & \cellcolor{white!70}$0.661$ & \cellcolor{white!70}$0.628$ \\
\midrule
\multicolumn{2}{c|}{$\Delta_{\mathrm{max - min}}$} & \textcolor{red}{$\downarrow0.472$} & \textcolor{red}{$\downarrow0.565$} & \textcolor{red}{$\downarrow0.568$} & \textcolor{red}{$\downarrow0.560$} & \textcolor{red}{$\downarrow0.413$} & \textcolor{red}{$\downarrow0.385$} \\
\bottomrule
\end{tabular}%
}
\end{center}
\caption{Recall at different context windows for JinaAI and MpNET, respectively. }
\label{tab:recall_table}
\vspace{-3mm}
\end{table}

\begin{table}[!ht]
\small
\begin{center}
\resizebox{0.85\columnwidth}{!}{%
\begin{tabular}{l|cccccc}
\toprule
\textbf{Type of Noise / Context Size} & \textbf{2\textit{k}} & \textbf{4\textit{k}} & \textbf{8\textit{k}} & \textbf{10\textit{k}} & \textbf{12\textit{k}} & \textbf{Avg.} \\
\midrule
Synthetic Distractors (S) & $0.73$ & $0.72$ & $0.71$ & $0.72$ & $0.70$ & $0.72$ \\
Natural Distractors (N) & $0.75$ & $0.75$ & $0.73$ & $0.72$ & $0.73$ & $0.74$ \\
Random Noise (N) & $0.75$ & $0.76$ & $0.75$ & $0.75$ & $0.74$ & $0.75$ \\
\midrule
N - S & \textcolor{ForestGreen}{$\uparrow0.01$} & \textcolor{ForestGreen}{$\uparrow0.03$} & \textcolor{ForestGreen}{$\uparrow0.02$} & \textcolor{black}{$0.00$} & \textcolor{ForestGreen}{$\uparrow0.02$} & \textcolor{ForestGreen}{$\uparrow0.02$} \\
R - S & \textcolor{ForestGreen}{$\uparrow0.02$} & \textcolor{ForestGreen}{$\uparrow0.04$} & \textcolor{ForestGreen}{$\uparrow0.04$} & \textcolor{ForestGreen}{$\uparrow0.03$} & \textcolor{ForestGreen}{$\uparrow0.04$} & \textcolor{ForestGreen}{$\uparrow0.04$} \\
\bottomrule
\end{tabular}%
}
\end{center}
\caption{Performance comparison on using three types of distractors, namely synthetic, natural, and random noise. Results show that synthetic distractors confuse the model more than natural distractors or random noise.}
\label{tab:noise_comparison}
\vspace{-8mm}
\end{table}

\paragraph{\textbf{Multilingual performance drops sharply with increasing linguistic distance from English}}

We observe that English consistently achieves the highest performance across both short and long context sizes. For long-context evaluation, we compare results at a $4k$ context window with a representative measure of extended context performance, calculated as the mean accuracy across all context lengths $\geq$$32k$. As shown in Table \ref{tab:LS-RQ-1}, performance is significantly lower for Chinese prompts, indicating a substantial disparity across languages. Moreover, there is a consistent decline in reasoning performance as the linguistic distance from English increases, particularly for languages with distinct phonetic and syntactic characteristics. Overall, performance across all languages in the \dataset\ benchmark remains relatively low, highlighting the challenges multilingual models face in solving complex tasks under long-context conditions.

\begin{figure}[!t]
\centering
\includegraphics[width=\linewidth]{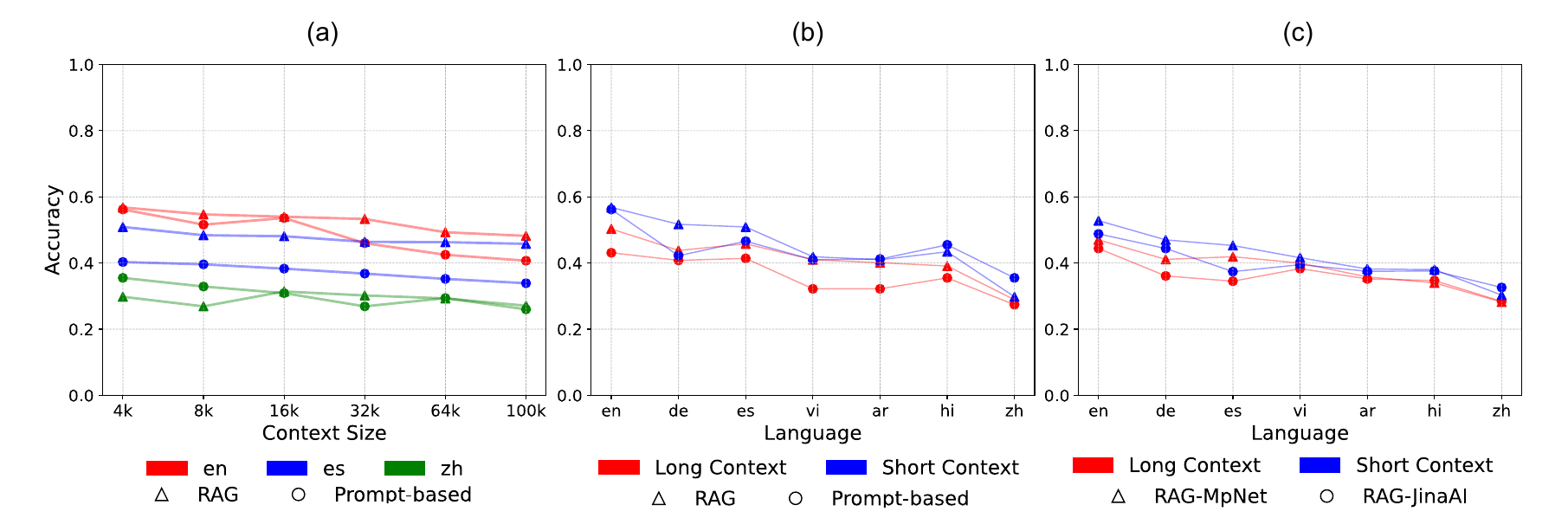}
    \caption{Performance comparison across different languages: (a) Performance comparison of prompt-based and RAG methods for best-performing language (en), second-best performing language (es) and worst-performing language (zh). (b) Performance comparison of prompt-based and RAG methods for different languages across short context length ($4k$) and longer context lengths ($\geq$ $32k$). (c) Performance of two RAG methods across short context length ($4k$) and longer context lengths ($\geq$ $32k$).}
    \label{fig:LS-RQ}
    \vspace{-6mm}
\end{figure}

To explain this trend further, we compare English with the second-best performing language (Spanish) and the lowest-performing (Chinese), as shown in Figure \ref{fig:LS-RQ}(a). We notice a clear decline in accuracy as context size increases, especially for prompt-based methods. Interestingly, when applying RAG methods, the performance gap between English and Spanish narrows consistently across all context sizes. Figure \ref{fig:LS-RQ}(b) further confirms that performance degradation is more pronounced in languages that are linguistically distant from English, and this decline is observed in both short- and long-context settings. Notably, the rate of degradation appears to follow a roughly linear relationship with linguistic distance. Lastly, Figure \ref{fig:LS-RQ}(c) compares the performance of different RAG models across languages. RAG-MpNet consistently outperforms RAG-JinaAI in both short- and long-context scenarios, with a more pronounced advantage in languages that are closer to English. However, for linguistically distant languages such as Arabic and Chinese, both RAG models perform similarly, indicating a shared limitation in handling these cases effectively.

\subsection{Task-wise Results}

\begin{table}[!t]
\small
\begin{center}
\resizebox{0.8\columnwidth}{!}{%
\begin{tabular}{lcccccccc}
\toprule
\multirow{8}{*}{\textbf{Short Context}} & &\textbf{RAG-JinaAI} &\textbf{RAG-MpNet} &\textbf{ZS} &\textbf{ICT} &\textbf{CoT} &\textbf{FS}\\
\cmidrule(lr){2-8}
& \textbf{en} & 0.526 & 0.573 & 0.536 & 0.540 & 0.584 & 0.492\\
& \textbf{de} & 0.492 & 0.515 & 0.481 & 0.494 & 0.470 & 0.529\\
& \textbf{es} & 0.415 & 0.500 & 0.459 & 0.466 & 0.545 & 0.430\\
& \textbf{vi} & 0.416 & 0.435 & 0.378 & 0.391 & 0.464 & 0.403\\
& \textbf{hi} & 0.428 & 0.433 & 0.423 & 0.417 & 0.452 & 0.425\\
& \textbf{ar} & 0.410 & 0.421 & 0.367 & 0.386 & 0.422 & 0.413\\
& \textbf{zh} & 0.336 & 0.305 & 0.238 & 0.238 & 0.431 & 0.416\\
\cmidrule(lr){2-8}
& \textbf{Avg.} & 0.432 & \textbf{0.455} & 0.412 & 0.419 & \underline{0.481} & 0.444\\
\multirow{8}{*}{\textbf{Long Context}} & & & & & & & \\
\cmidrule(lr){2-8}
& \textbf{en} & 0.480 & 0.514 & 0.434 & 0.437 & 0.418 & 0.434\\
& \textbf{de} & 0.410 & 0.465 & 0.432 & 0.444 & 0.327 & 0.430\\
& \textbf{es} & 0.382 & 0.473 & 0.419 & 0.410 & 0.386 & 0.457\\
& \textbf{vi} & 0.404 & 0.415 & 0.302 & 0.326 & 0.331 & 0.329\\
& \textbf{hi} & 0.392 & 0.390 & 0.366 & 0.359 & 0.324 & 0.371\\
& \textbf{ar} & 0.398 & 0.404 & 0.321 & 0.327 & 0.320 & 0.321\\
& \textbf{zh} & 0.294 & 0.283 & 0.240 & 0.220 & 0.297 & 0.283\\
\cmidrule(lr){2-8}
& \textbf{Avg.} & 0.394 & 0.421 & 0.359 & 0.360 & 0.343 & 0.375\\
\bottomrule
\end{tabular}}
\caption{Performance of different prompt-based and RAG methods for different context lengths. "Short Context" refers to context sizes ($4k$-$16k$), and "Long Context" refers to context sizes ($\geq$ $32k$). For short context lengths, we observe that RAG-MpNet performs the best among the RAG methods, and cot performs the best among the prompt-based methods. For long context lengths, we observe that RAG-MpNet performs the best among the RAG methods and few-shot performs the best among the prompt-based methods}
\label{tab: RS-RQ}
\vspace{-8mm}
\end{center}
\end{table}

Heatmaps in Figures \ref{fig:result_table5a} and \ref{fig:result_table5b} show the task-wise performance of selected methods across four context windows -- $16k$, $32k$, $64k$, and $128k$, respectively. The colour patterns across both figures reveal a clear disparity in task difficulty, with some tasks showing significantly lower accuracy than others, regardless of the language. Among these, Basic Factoid QA achieves the highest accuracy across all languages and context lengths. This suggests that the model is generally effective at retrieving isolated facts, even in long contexts. Yes/No (Negation) questions also perform relatively well, though results vary more across languages. In contrast, tasks that require multi-step reasoning, such as Fact Chaining and Argument Relations, show noticeably low accuracy. This highlights the model’s limitations in maintaining coherent reasoning over long contexts. Similarly, Aggregation tasks, such as Counting and Lists/Sets, show only moderate performance, with a significant decline in low-resource and morphologically complex languages. The most pronounced decline is observed in questions on Indefinite Knowledge, which tests a model’s ability to handle ambiguity or incomplete information. Across all languages, this task consistently shows low accuracy, suggesting the challenges in handling epistemic reasoning and uncertainty, especially in longer contexts.

\begin{table}[!t]
\small
\begin{center}
\resizebox{0.6\columnwidth}{!}{%
\begin{tabular}{lccccc}\toprule
&\textbf{Context size} & \shortstack{\textbf{Prompt based} \\ \textbf{methods}} & \shortstack{\textbf{RAG} \\ \textbf{methods}} &\textbf{Average} \\\midrule
\multirow{2}{*}{\textbf{en}} &$4k$ &0.562 &0.568 &0.565 \\
&\shortstack{$\geq$ $32k$} &0.431 &0.503 &\textbf{0.467} \\
\midrule
\multirow{2}{*}{\textbf{de}} &$4k$ &0.422 &0.517 &0.470 \\
&\shortstack{$\geq$ $32k$} &0.408 &0.438 &0.423 \\
\midrule
\multirow{2}{*}{\textbf{es}} &$4k$ &0.466 &0.509 &0.488 \\
&\shortstack{$\geq$ $32k$} &0.414 &0.458 &0.436 \\
\midrule
\multirow{2}{*}{\textbf{hi}} &$4k$ &0.455 &0.434 &0.444 \\
&\shortstack{$\geq$ $32k$} &0.355 &0.391 &0.373 \\
\midrule
\multirow{2}{*}{\textbf{ar}} &$4k$ &0.411 &0.410 &0.411 \\
&\shortstack{$\geq$ $32k$} &0.322 &0.401 &0.362 \\
\midrule
\multirow{2}{*}{\textbf{zh}} &$4k$ &0.355 &0.298 &0.327 \\
&\shortstack{$\geq$ $32k$} &0.274 &0.289 &\underline{0.281} \\
\midrule
\multirow{2}{*}{\textbf{vi}} &$4k$ &0.410 &0.419 &0.414 \\
&\shortstack{$\geq$ $32k$} &0.322 &0.410 &0.366 \\
\bottomrule
\end{tabular}
}
\end{center}
\caption{Performance of prompt-based and RAG methods for $4k$ and for longer context lengths ($\geq$ $32k$) for different languages.}
\label{tab:LS-RQ-1}
\vspace{-6mm}
\end{table}

\begin{figure}
    \centering
    \includegraphics[width=\linewidth]{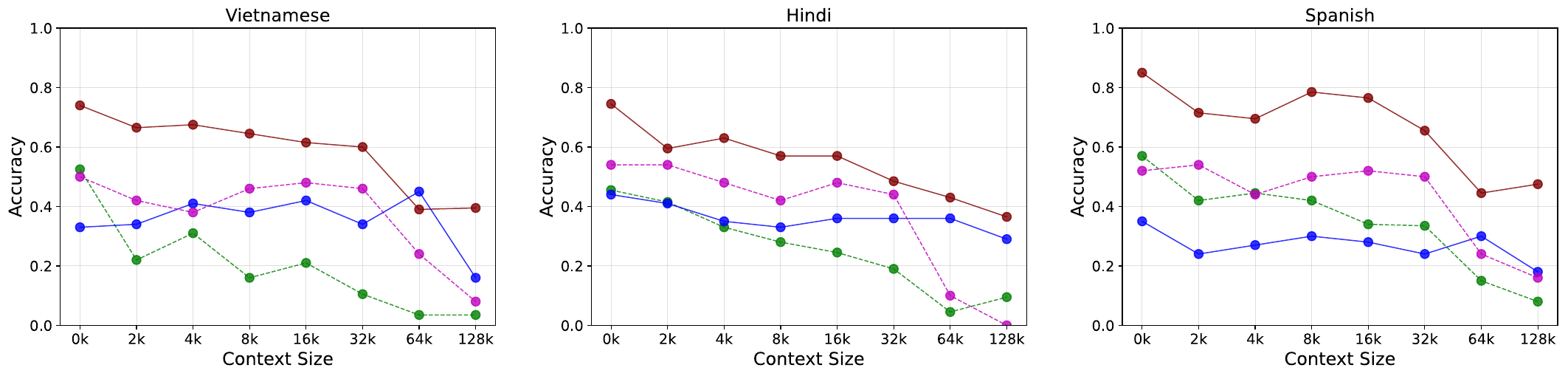}
    \includegraphics[width=\linewidth]{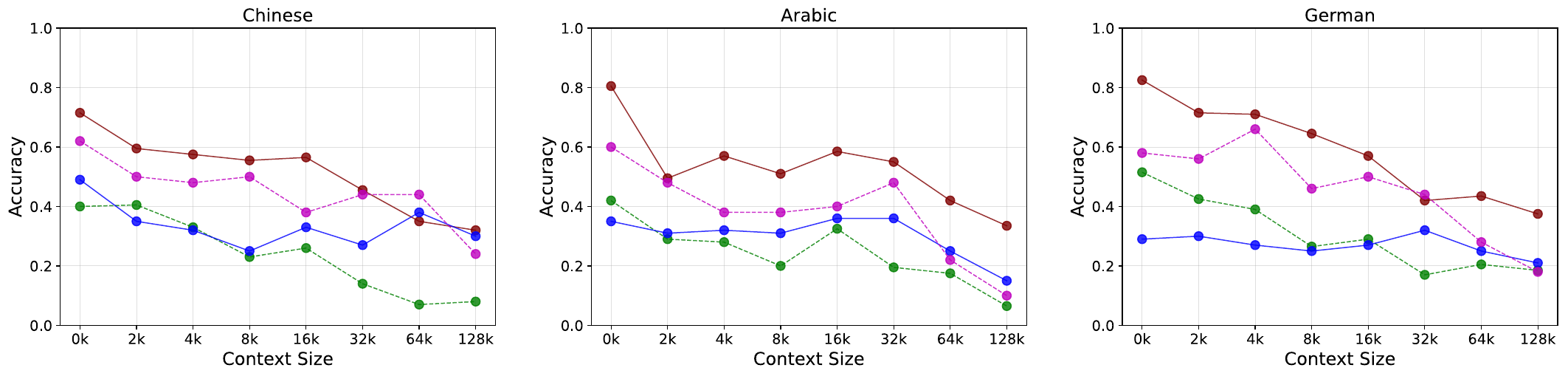}

    \includegraphics[width=1.0\linewidth]{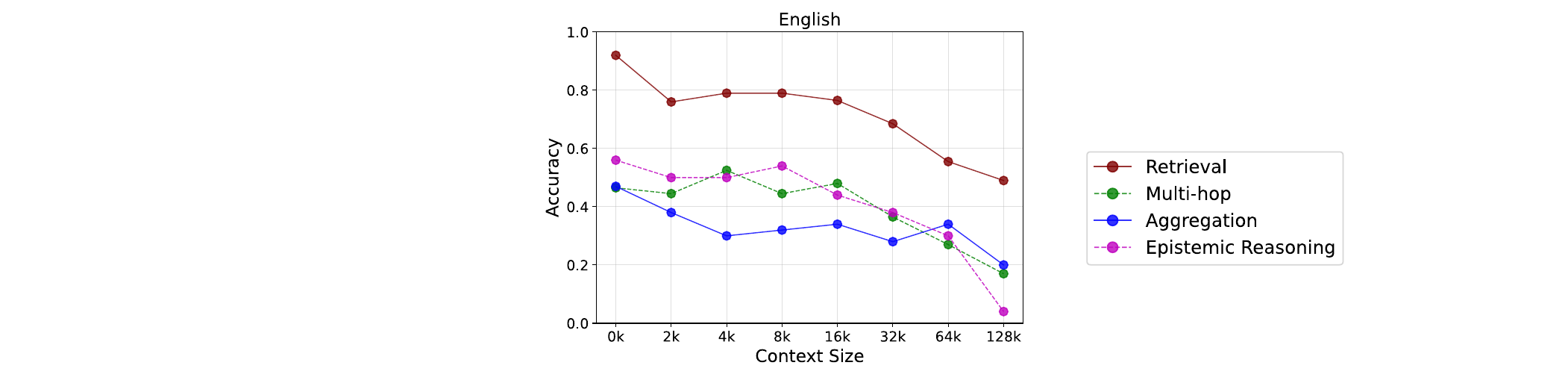}

    \caption{Task-wise performance of Llama-3.1-Instruct on selected languages. We group the tasks into four categories as discussed in Section \ref{sec:task_categories}.
    } 
    \label{fig:task-specific-results}
    \vspace{-8mm}
\end{figure}

\paragraph{\textbf{Retrieval versus reasoning}}  In Figure \ref{fig:task-specific-results}, we plot results for the four task categories --  Retrieval, Multi-hop Inference, Aggregation, and Uncertainty, respectively. Regardless of language, the central observation remains that there is a visible demarcation between Retrieval and other categories throughout the increasing context lengths. The performance of the model in retrieval is the highest accuracy in languages and context lengths, suggesting that models can effectively locate and recall isolated information. Aggregation slightly outperforms Multi-hop Inference, although both tasks present similar challenges due to their dependence on integrating multiple facts from the context. The most difficult category is Uncertainty, which involves epistemic reasoning and tests the model's ability to recognize ambiguous or incomplete information. Performance in this category is consistently the lowest across all settings. Instead of expressing uncertainty or deferring judgment when necessary, the model tends to provide overconfident and incorrect responses. 

In summary, our results show that while LLMs are effective at retrieving isolated facts, their capabilities degrade substantially when required to reason over long contexts. Moreover, the  
stark gap in accuracy between retrieval and reasoning tasks warrants new evaluation setups like \dataset\ that extend beyond surface-level retrieval or recall.

\subsection{Effective Context Size}
Table \ref{tab:main_results} summarizes the {effective context size} for all baseline methods. We define the effective context window (ECW) as the maximum context length at which the model’s accuracy stays within 30\% of its performance without distractors. Our results show that the effective context size is approximately $16k$ tokens for the best prompt-only and $32k$ tokens for the best RAG method. This suggests that the model is able to effectively use only up to 25\% to 30\% of its advertised context length. Thus, despite large claimed context windows, models may be failing to fully utilize them for reasoning tasks, particularly for multilingual inputs. Given that open-weight LLMs are increasingly adopted in high-stakes applications, this limitation becomes an important consideration.

\subsection{Ablation Study}
\begin{figure}
    \centering
    \includegraphics[width=\linewidth]{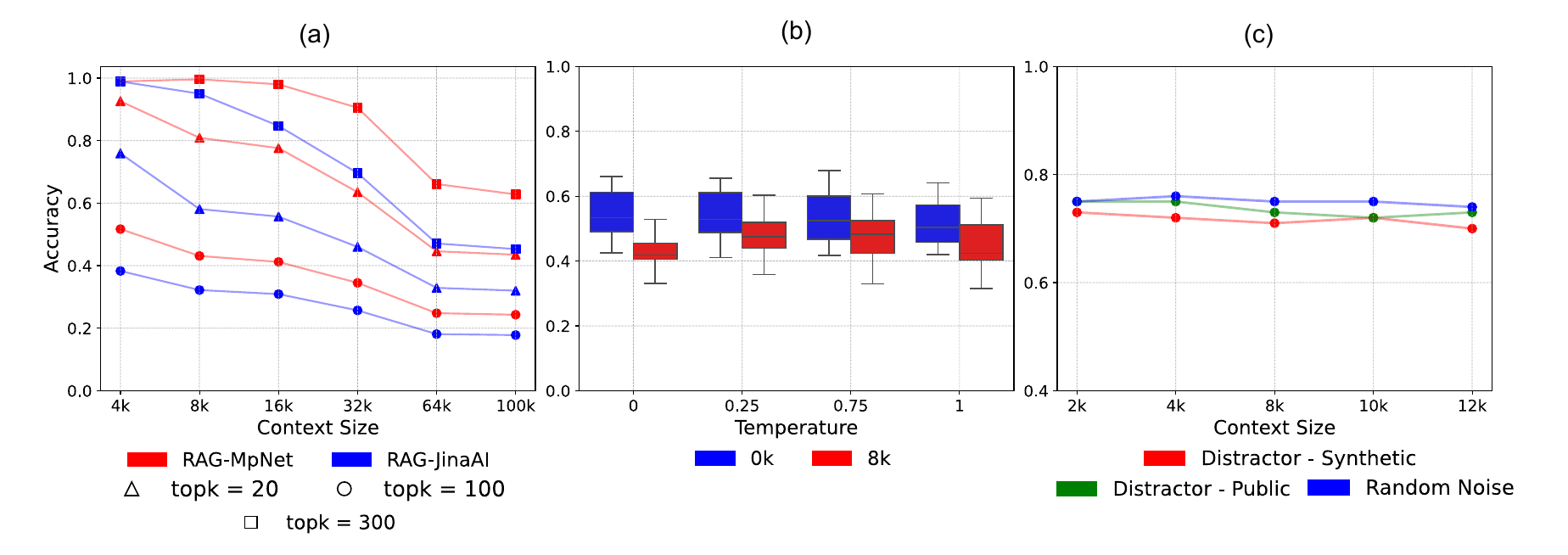}
    \caption{Ablation studies: (a) Performance of two different RAG methods with different numbers of retrieved documents. (b) Effect of temperature scaling at 0k context size and 8k context size. (c) Performance across different context sizes is affected by the type of distractors used to increase the context size.}
    \label{fig:ablations}
    \vspace{-5mm}
\end{figure}

We conduct a series of ablation experiments to better understand the factors influencing model performance under multilingual long-context settings.

\paragraph{\textbf{Effect of top-k retrieval}}
We evaluate the impact of the number of retrieved sentences in the RAG pipeline by varying the \textit{top-k} values ($20$, $100$, and $300$). As shown in Figure \ref{fig:ablations}(a), increasing the number of retrieved sentences improves performance across context sizes. For both JinaAI and MpNet, the gain in performance is particularly notable at higher context lengths. This suggests that using RAG may counteract information dispersion in long inputs, although to a limited extent. 

\paragraph{\textbf{Effect of distractor type}} 
As defined in Section \ref{sec:distractors}, we compare three types of distractors used to inflate context length: synthetic, natural, and random noise. As shown in Figure \ref{fig:ablations}(c), synthetic distractors lead to slightly lower performance than the other types. This suggests that in-distribution distractors, which are structurally similar to relevant content, make the task harder compared to other distractors. Thus, using synthetic distractors is the closest to reflecting real-world difficulty.

\paragraph{\textbf{Effect of Temperature Sampling}} 

To evaluate the impact of the decoding strategy, we compared model performance using temperature sampling at four values: $0$, $0.25$, $0.5$, and $1$, respectively. As shown in Figure \ref{fig:ablations}(b), varying the temperature leads to minimal changes in accuracy, indicating that model outputs remain stable across different sampling conditions. This suggests that performance is not significantly influenced by the decoding strategy. Combined with our statistical significance tests (Section \ref{sec:stat_tests}), which demonstrate performance convergence beyond $250$ evaluation samples, these findings reinforce the overall reliability and robustness of our experimental results.

\subsection{Statistical Tests}
\label{sec:stat_tests}

\begin{figure*}[t]
\centering

\begin{subfigure}{0.48\textwidth}
  \centering
  \includegraphics[width=\linewidth]{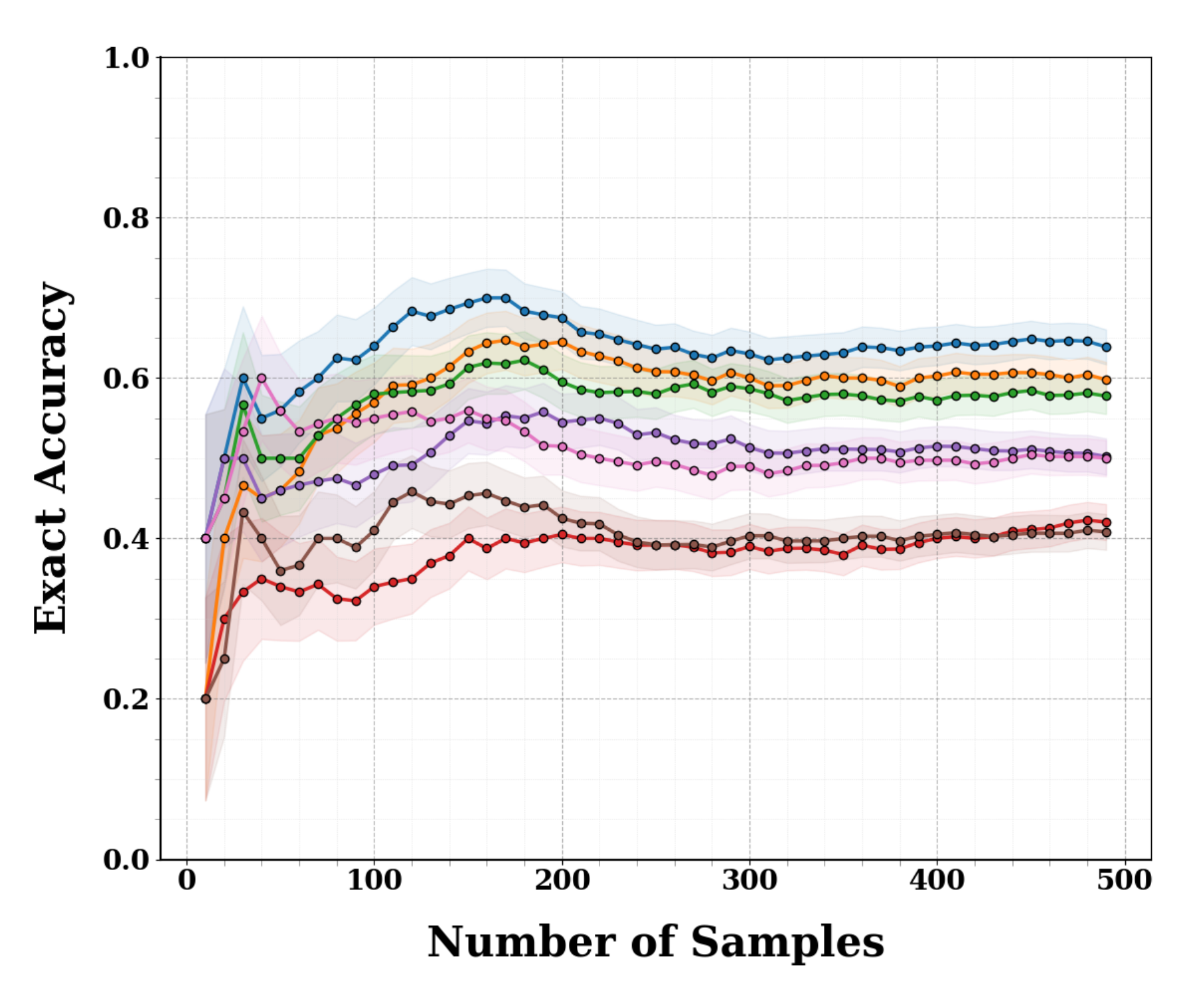}
  \caption{Mean and standard error for selected languages at baseline (no context)}
  \label{fig:top_left}
\end{subfigure}
\hfill
\begin{subfigure}{0.48\textwidth}
  \centering
  \includegraphics[width=\linewidth]{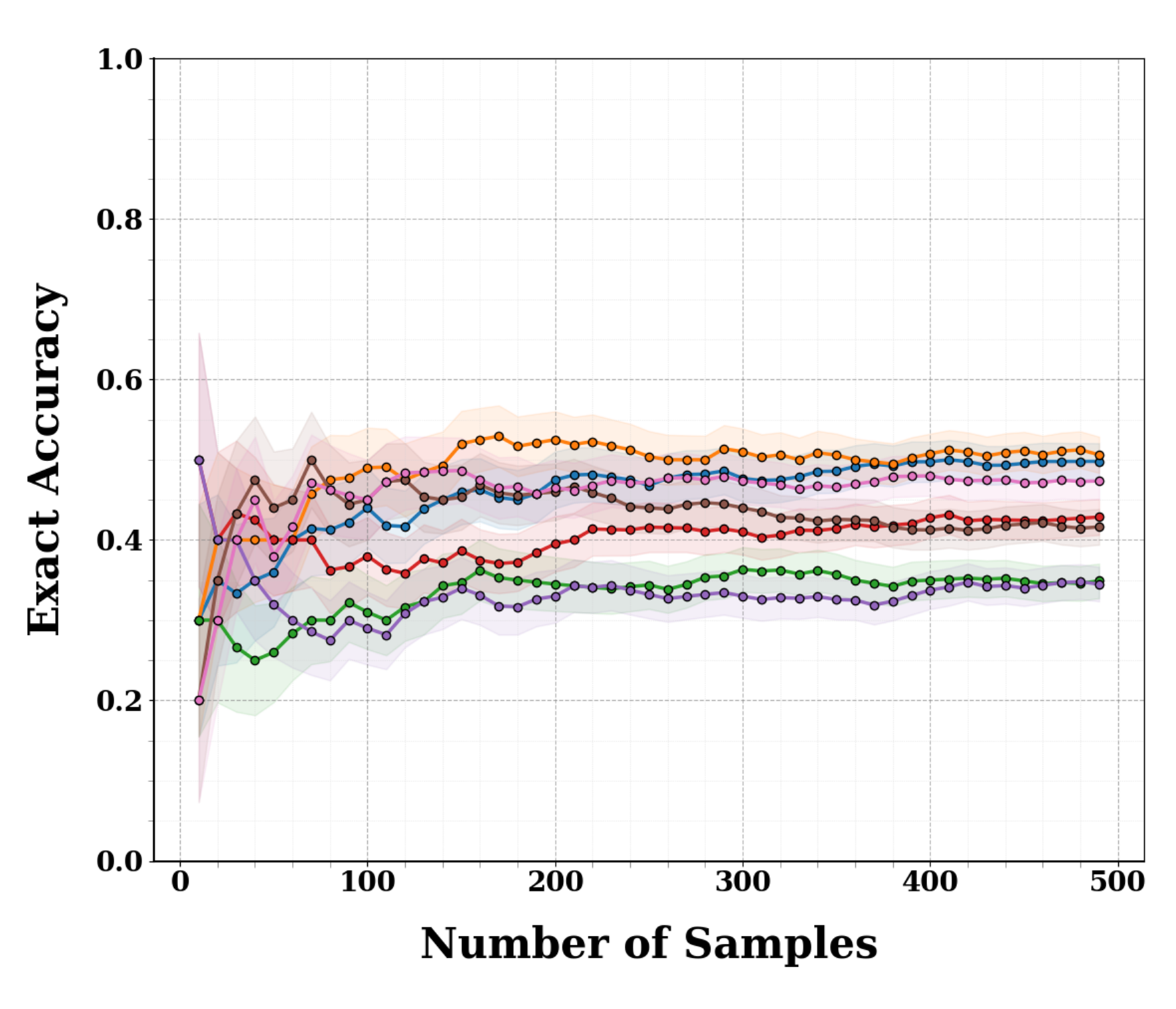}
  \caption{Mean and standard error for selected languages at short context ($4k$)}
  \label{fig:top_right}
\end{subfigure}


\begin{subfigure}{0.48\textwidth}
  \centering
  \includegraphics[width=\linewidth]{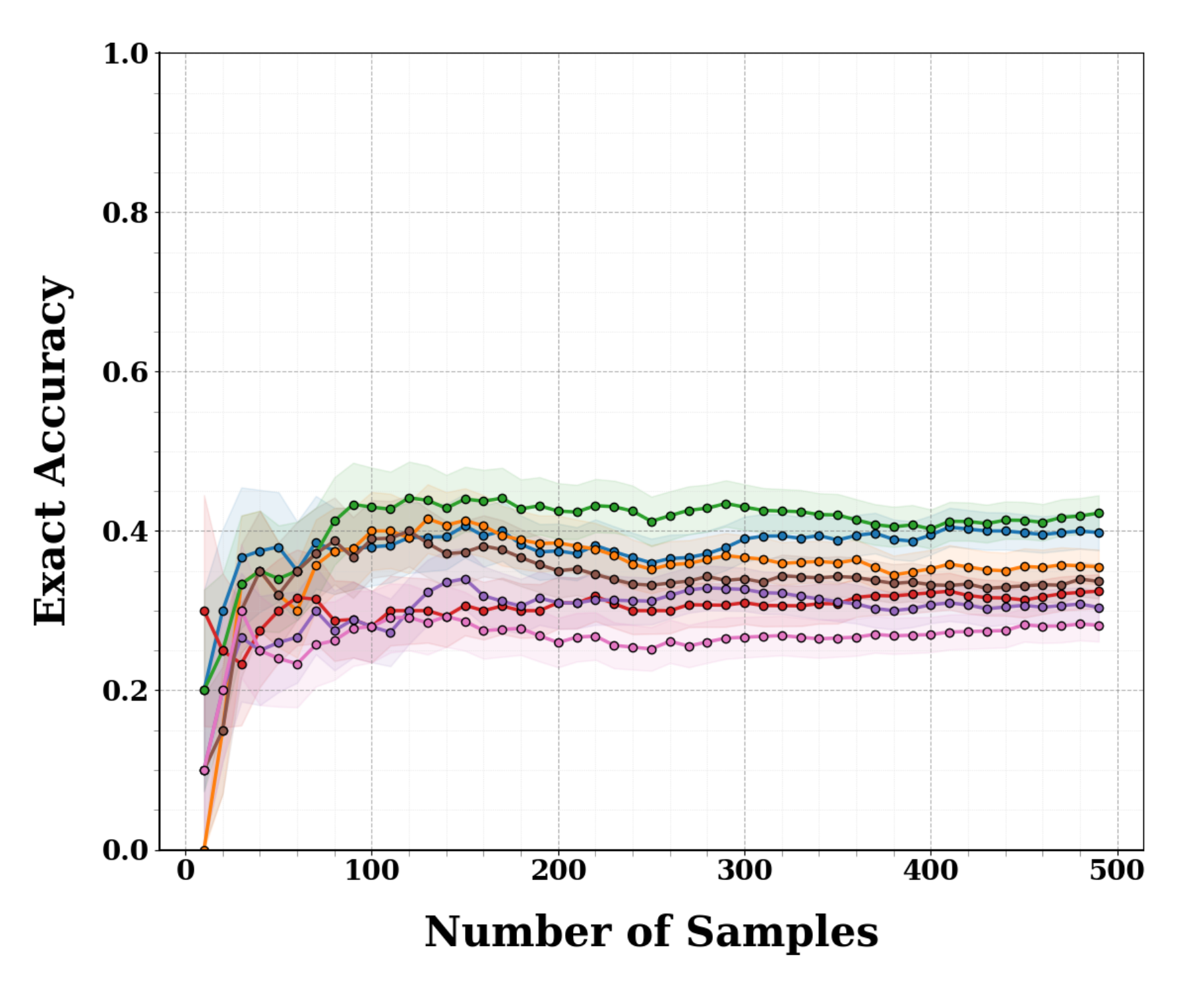}
  \caption{Mean and standard error for selected languages at long context ($32k$)}
  \label{fig:bottom_left}
\end{subfigure}
\hfill
\begin{subfigure}{0.48\textwidth}
  \centering
  \includegraphics[width=\linewidth]{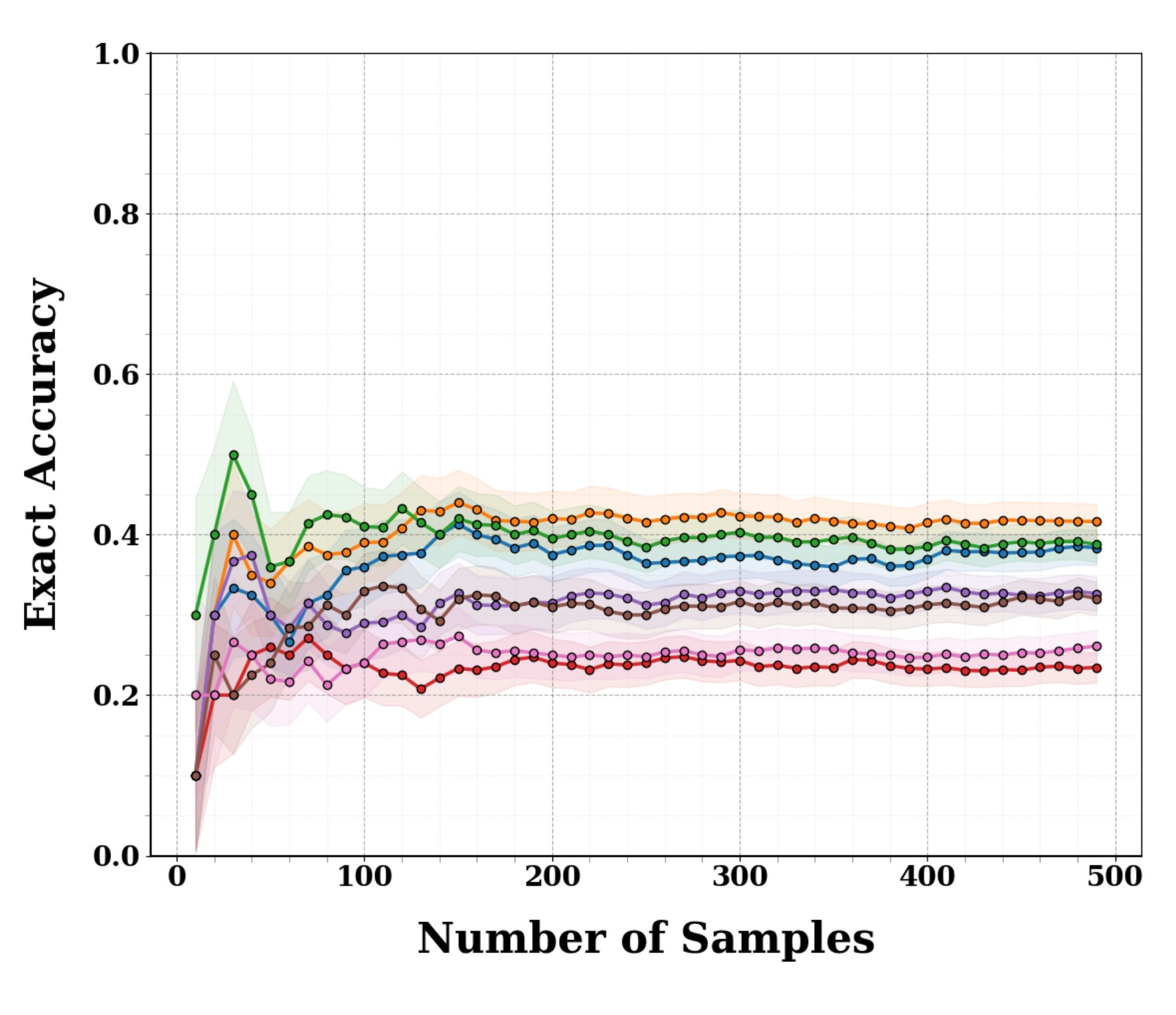}
  \caption{Mean and standard error for selected languages at long context ($128k$)}
  \label{fig:bottom_right}
\end{subfigure}

\caption{Exact accuracy of Llama 3.1-Instruct after varying the sample size in the evaluation (test) set. Solid lines represent accuracy, while shaded areas indicate the standard error. Legend (by color): en (blue), de (orange), es (green), zh (red), vi (purple), hi (brown), ar (pink).}
\label{fig:stat_test}
\vspace{-8mm}
\end{figure*}

We conduct binomial significance tests on our main results for \text{Llama3.1-Instruct}. Figure \ref{fig:stat_test} shows the significance tests across four settings: baseline, short-context, and long-context, respectively. These tests are performed over evaluation (test) sets of increasing size, ranging from $10$ to $500$ samples.  

We find that across all selected languages, the exact accuracy begins to stabilize between $250$ and $500$ samples. Additionally, the standard error decreases significantly as the number of evaluation samples exceeds $250$. This indicates that using $250$ or more samples is sufficient to yield reliable and consistent evaluation results. Since our main experiments are conducted using $500$ samples per language, we can confidently state that our results are statistically significant. We detail our hypothesis testing approach below.

Let \(\text{Binomial}(1, p)\) represent the binomial distribution, where \(p\) is the probability of success on an individual trial. The standard error ($SE$) is computed using the formula:
\begin{equation}
\begin{aligned}
SE = \sqrt{\frac{p(1 - p)}{s}},
\end{aligned}
\end{equation}
where \(s\) is the number of trials (i.e., evaluation samples). We vary \(s\) from $10$ to $500$ in increments of $10$, i.e., $20$, $30$, $40$, and so on — randomly sampling instances from \dataset\ for each selected language at every step.
\section{Related Work}

\paragraph{\textbf{Multilingual reasoning}} The world is inherently multilingual, and for LLMs to be truly deployable in global-facing applications, they must be capable of thinking, understanding, reasoning, and responding across a broad spectrum of languages \cite{shi2022languagemodelsmultilingualchainofthought, zhao-largelanguagemodelshandle-2024, cao2024mindtonguesdeepdive}. In recent years, there has been steady progress in enhancing LLMs’ multilingual reasoning abilities, enabled by advancements in in-context prompting \cite{Tanwar2023MultilingualLA}, multi-step instruction following \cite{multi-step-instruction-following}, and pre-translation methods \cite{ICT0}. A key driver of this progress has been the development of reliable benchmarks like MGSM \cite{mgsm}, XCOPA \cite{xcopa}, and XL-WiC \cite{raganato-etal-2020-xl}. These allow researchers to evaluate various aspects of reasoning, such as arithmetic reasoning, commonsense understanding, and semantic inference. However, current work in multilingual reasoning focuses on short, isolated prompts and does not evaluate LLMs in extended, multilingual contexts. Our work builds on this line of inquiry by investigating how LLMs behave under extended multilingual contexts, a setting which is an increasingly common scenario in real-world applications where the input can span across multiple sources, dialogues, or documents \cite{application-Lee_2022, application-rozière2024codellamaopenfoundation, application-shah2024multidocumentfinancialquestionanswering}.

\paragraph{\textbf{Long-context evaluation}}
Currently, long-context evaluation is dominated by the popular \textit{Needle-in-a-Haystack (NIAH)} tests \cite{gregory2023-NIAH}, where an LLM is tasked with finding one or more pieces of information (the \textit{"needle"}) from long, irrelevant background texts (the \textit{"haystack"}). Existing multilingual long-context evaluation benchmarks have extended the NIAH test to multiple languages \cite{mLongRR, hengle2024_multilingualneedlehaystackinvestigating, kim2025_rulermeasureallbenchmarking, huang2025_benchmaxcomprehensivemultilingualevaluation}. While these are suitable for evaluating surface-level \textit{retrieval} or \textit{recall}, they are not sufficient to evaluate multilingual reasoning under long-context settings \cite{goldman2024reallylongcontextneed}. Subsequently, a growing body of research has started to look beyond the retrieval-focused NIAH tests for long-context evaluation, albeit limited to monolingual English settings \cite{kuratov2024babilongtestinglimitsllms, michelangelo, karpinska2024-Thousand-and-one-pairs}. For instance, \citet{michelangelo} proposed the Latent Structure Queries (LSQ) framework, which necessitates the LLM to understand the \textit{latent information} within extended contexts. More recently, \citet{kuratov2024babilongtestinglimitsllms} proposed multiple synthetic reasoning tasks by extending the bAbI dataset to long-context settings. We take inspiration from them while curating \dataset.

\paragraph{\textbf{Synthetic and realistic benchmarks}} Creating realistic benchmarks for long-context evaluation is often challenging due to factors like increased complexity, limited resources, and the high cost of human annotation \cite{hsieh2024-Ruler}. As a result, recent research has increasingly turned to synthetically generated datasets as proxies for real-world performance \cite{michelangelo}. Because synthetic datasets are easier to construct and evaluate, most existing long-context benchmarks are synthetic, commonly following the NIAH framework \cite{gregory2023-NIAH, hsieh2024-Ruler, An2023-LEvalIS, bai-etal-Longbench-2024, kim2025_rulermeasureallbenchmarking}. In contrast, realistic benchmarks typically center on tasks that reflect practical applications, such as summarization \cite{kim2024fablesevaluatingfaithfulnesscontent} and question answering \cite{kuratov2024babilongtestinglimitsllms, shaham2023zeroscrollszeroshotbenchmarklong, hengle2024_multilingualneedlehaystackinvestigating}. While synthetic benchmarks offer greater control and flexibility, realistic benchmarks are more reliable for assessing real-world performance. \dataset\ brings together the strengths of both approaches -- it builds on the synthetic bAbI dataset while presenting tasks in the SQuAD format \cite{rajpurkar-etal-2016-SQUAD}, aligning with the structure and demands of real-world applications.
\section{Conclusions}
This work introduces \dataset, a synthetic benchmark to assess LLMs' reasoning capabilities over multilingual long contexts. \dataset\ offers a rigorous evaluation framework with parallel prompts across seven languages and multiple context lengths. Unlike existing multilingual long-context benchmarks, \dataset\ is explicitly designed to resist evaluation leakage and short-circuiting. Going beyond the popular \textit{needle-in-a-haystack} test, \dataset\ incorporates tasks that require multi-hop inference, aggregation, and epistemic reasoning. Through extensive experiments on an open-weight LLM, we observe a clear performance disparity between high- and low-resource languages, especially in tasks involving multi-step reasoning and uncertainty resolution. While off-the-shelf RAG methods offer some improvement in extending the \textit{effective context length}, they do not fully overcome the challenges associated with reasoning over extremely long contexts. Furthermore, we observe that performance is notably skewed towards retrieval-based tasks, with a considerable drop in accuracy for other task categories like epistemic reasoning. This suggests that current benchmarks, which focus primarily on surface-level retrieval, may significantly overestimate the true long-context capabilities of LLMs. In summary, our findings highlight that multilingual long-context reasoning remains an open and unsolved problem. Addressing this is crucial if we are to tackle long-term and long-tail challenges, especially when it comes to deploying these LLMs in global, high-stakes applications.
\section{Limitations and Ethical Statement}
\label{sec:limitations}
Recent API-based models such as GPT-4 \cite{gpt4-technical-report} and Claude \cite{claude3opus} have shown impressive reasoning capabilities. However, due to their high cost — particularly given the large size of our evaluation prompts — we were unable to include them in our experiments. While we did experiment with a lightweight re-ranking model (JinaAI), future work could explore more complex and dedicated RAG pipelines, such as GraphRAG \cite{edge2025localglobalgraphrag}. We chose not to pursue this direction in the current work, as our goal was to evaluate retrieval using simple, off-the-shelf methods that introduce minimal computational overhead.

An interesting extension of \dataset\ would be cross-lingual evaluation, where the question, passage, or background texts appear in different languages. Although \dataset\ supports such experiments, this was outside the scope of our current study. Another promising direction for future work is a deeper interpretability analysis of model behavior in long-context settings, for instance, comparing retrieval versus reasoning performance or analyzing how models process and attend to information across long contexts. This could involve techniques from mechanistic interpretability or studying self-attention patterns, similar to recent work like \cite{hsieh-found-in-the-middle-2024}.

Finally, \dataset\ is fully synthetic and did not involve any human annotators during its construction. As a result, this work does not require additional ethical review from an institutional review board. All models, datasets, and scientific artefacts used in this study have been properly cited where applicable.

\clearpage
\newpage
\appendix
\label{sec:appendix}

\appendixsection{Translation Process}
\label{appendix:translation}

\vspace{-4mm}

\begin{figure}[t]
\centering \includegraphics[width=0.65\columnwidth]{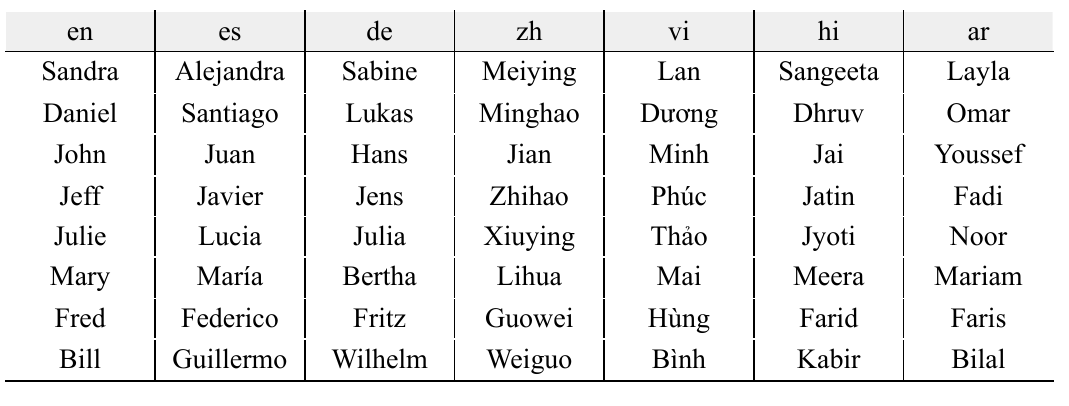}
\small
\caption{{We replace the English bAbI characters with a culturally appropriate name in
each of the target languages before translation.}} 
\label{tab:crosslingual_names}
\vspace{-6mm}
\end{figure}

\begin{table}[!ht]
\begin{center}
\begin{minipage}[t]{0.48\linewidth}
\centering
\resizebox{\linewidth}{!}{%
\begin{tabular}{l|cccccc}
\toprule
\textbf{Dataset Instance} & \textbf{es} & \textbf{de} & \textbf{zh} & \textbf{vi} & \textbf{hi} & \textbf{ar} \\
\midrule
passages  & $0.89$ & $0.90$ & $0.85$ & $0.88$ & $0.84$ & $0.85$ \\
questions & $0.89$ & $0.91$ & $0.87$ & $0.88$ & $0.87$ & $0.87$ \\
answers   & $0.89$ & $0.89$ & $0.87$ & $0.87$ & $0.87$ & $0.87$ \\
\bottomrule
\end{tabular}%
}
\caption*{\small \textbf{(A) Translation similarity}}
\end{minipage}
\hfill
\begin{minipage}[t]{0.48\linewidth}
\centering
\resizebox{\linewidth}{!}{%
\begin{tabular}{l|cccccc}
\toprule
\textbf{Dataset Instance} & \textbf{es} & \textbf{de} & \textbf{zh} & \textbf{vi} & \textbf{hi} & \textbf{ar} \\
\midrule
passages  & $0.98$ & $0.98$ & $0.98$ & $0.97$ & $0.98$ & $0.98$ \\
questions & $0.98$ & $0.98$ & $0.98$ & $0.99$ & $0.99$ & $0.98$ \\
answers   & $0.97$ & $0.97$ & $0.97$ & $0.94$ & $0.96$ & $0.96$ \\
\bottomrule
\end{tabular}%
}
\caption*{\small \textbf{(B) Back-translation similarity}}
\end{minipage}
\caption{{(A) Semantic similarity between source (English) and target languages. We find that source and target texts have high semantic alignment for all languages. (B) The semantic similarity between the source (English) and back-translated texts from the target language. We observe a negligible error rate (high semantic similarity) between the source and back-translated texts for all languages. Results from (A) and (B) prove the robustness of our translation process.}}
\label{tab:translation_QE}
\end{center}
\end{table}


Our data curation process uses machine translation to extend the bAbI (English) dataset to selected languages. One might argue that this approach introduces risks such as translation drift or quality issues, which could affect benchmark reliability. An obvious alternative is to use human translators, which would offer higher reliability but is costly and difficult to scale. To understand why we opted against this, it is essential to consider the nature of bAbI dataset. As illustrated in Table~\ref{fig:dataset_example}, bAbI comprises discrete, atomic factual statements that follow a consistent subject–verb–object format. These are constructed from a closed set of entities, actions, and vocabulary items \cite{weston2015_babi}. Such a closed and formulaic structure substantially reduces the likelihood of translation errors arising from contextual ambiguity, co-reference, or idiomatic usage. Moreover, our translation process is applied at the sentence level, with each unit being an independent, context-free, and unambiguous factual statement \cite{weston2015_babi}. This further reduces the risk of any translation drift.  Furthermore, we use the Google Translate API \footnote{https://cloud.google.com/translate/docs/reference/api-overview - Google Translate API}, which is one of the most high-performing and widely used machine translation tools \footnote{\href{https://docs.google.com/spreadsheets/d/1fJQLMj8O5z3Q7eKDxi1tNNrFipiEL0UDyaEF0fleZ54/htmlview?pli=1}{Performance report of Google Translation API}}. We assess the quality of the translated data in \dataset\ by conducting a translation quality estimation (QE) following prior work \cite{automatic-mt-evaluation}~\footnote{We compute semantic-similarity between source and translated texts using XLM-Roberta-Large.}. As shown in Table \ref{tab:translation_QE} (A), we find high semantic similarity between the English source texts and their translations across all languages. Additionally, the back-translation results in Table \ref{tab:translation_QE} (B) indicate that the average information loss during translation is below 2\%. We also manually reviewed a randomly selected 5\% subset of \dataset, and found no instances of misleading translations. Together, these findings support the quality and reliability of the translated data in \dataset. 


\appendixsection{Task Instructions}
\begin{table*}[!h]
\small
\begin{center}
\scalebox{0.9}{
  \renewcommand{\arraystretch}{1.5}
\begin{tabular}{
>{\raggedright\arraybackslash}p{0.5cm}|
>{\raggedright\arraybackslash}p{2cm}|
>{\raggedright\arraybackslash}p{2cm}|
>{\raggedright\arraybackslash}p{6.5cm}
}
\hline
\textbf{Task ID} & \textbf{Task Name} & \textbf{Total Test Samples (per language)} & \textbf{Prompt Instruction \texttt{\{Instruct\}} } \\
\hline
1 & Basic Factoid QA & 200 & Read the provided text, which contains facts about the locations of different individuals mixed with unrelated information. Answer the question based only on these facts. If a person appears in multiple locations, use their most recent location to determine the answer. \\
\hline
2 & Yes/No or Negation & 200 & Read the provided text, which contains facts mixed with unrelated information. Answer the question while accurately interpreting negation with either "Yes" or "No," based only on the provided facts. \\
\hline
3 & Fact Chaining & 200 & Read the provided text, which contains facts about different events mixed with unrelated information. Answer the question by identifying and combining multiple relevant facts. \\
\hline
4 & Association & 100 & Read the provided text, which contains facts about different relationships mixed with unrelated information. Answer the question based on relationships involving two or more entities. \\
\hline
5 & Counting & 100 & Read the provided text, which contains facts mixed with unrelated information. Answer the counting question by identifying and counting the relevant items, then provide the correct number. \\
\hline
6 & List / Set & 100 & Read the provided text, which contains facts mixed with unrelated information. Answer the question by identifying all relevant items and listing them in the correct format. \\
\hline
7 & Indefinite Knowledge & 100 & Read the provided text, which contains facts mixed with unrelated information. Answer the question based only on the given facts. If the necessary information is not explicitly provided, respond with "I don't know (IDK)". \\
\hline
\end{tabular}}
\caption{Prompt instructions and test sample distribution for each task type.}
\label{tab:task-instructions}
\end{center}
\end{table*}
\newpage

\appendixsection{Task Example}
\begin{table*}[!h]
\small
\begin{center}
\scalebox{0.9}{
  \renewcommand{\arraystretch}{1.5}
\begin{tabular}{
>{\raggedright\arraybackslash}p{2cm}|
>{\raggedright\arraybackslash}p{3.5cm}|
>{\raggedright\arraybackslash}p{3.5cm}|
>{\raggedright\arraybackslash}p{4cm}
}
\hline
\textbf{Task} & \textbf{Description} & \textbf{Example} & \textbf{Task Type} \\
\hline
Basic Factoid QA & The answer can be retrieved from one directly relevant fact (independent statement) within the context. & Passage: "Daniel went back to the bedroom." Q: "Where is Daniel?" A: \textcolor{ForestGreen}{"bedroom"} & \textbf{Retrieval (Needle in a Haystack)}. The model must locate one key sentence among possibly many. No inference or chaining.  \\
\hline
Yes/No or Negation & The answer is binary (yes/no), possibly involving negation or contradiction in the context. & Passage: "Mary is not in the kitchen. Mary is not in the garden." Q: "Is Mary in the garden?" A: \textcolor{ForestGreen}{"no"} & \textbf{Retrieval (Needle in a Haystack)}. The model needs to interpret a negated statement and confirm or deny the query. \\
\hline
Fact Chaining & The answer is derived by combining two or more facts across multiple sentences, often requiring tracking over time. & Passage: David bought a football. He forgot it in his bedroom \{...\} Mary gave it to Sandra \{...\} Sandra threw it away in the garden.  Q: "Where was the football before the garden?" A: \textcolor{ForestGreen}{"bedroom"} & \textbf{Multi-hop Reasoning}. The model is required to track a chain of actions across time --- for instance where the football moved, and what happened before a specific event. \\
\hline
Association & The answer is inferred through understanding spatial, logical, or relational connections between entities. & Passage: "The bedroom is north of the bathroom. The bedroom is south of the kitchen." Q: "What is the kitchen north of?" A: \textcolor{ForestGreen}{"bedroom"} & \textbf{Multi-hop Reasoning}. The model is required to understand and reason over relationships (e.g., directions) rather than just follow entity movements. \\
\hline
Counting & Requires counting the number of relevant entities or objects based on events in the Passage. & Passage: Apple is passed from Mary $\rightarrow$ Daniel $\rightarrow$ Mary $\rightarrow$ Sandra. Ball is passed from Mary $\rightarrow$ Sandra $\rightarrow$ Daniel. Q: "How many objects is Daniel carrying?" A: \textcolor{ForestGreen}{"one"} & \textbf{Aggregation}. The model is required to track state changes across multiple characters and actions. \\
\hline
List / Set & The answer is a set or list of entities retrieved from the context, often based on participation or possession. & Passage: Mary picked up the milk, grabbed a football, and dropped her purse in the office. After going home, she left the football and picked the cup from kitchen. Q: "What is Mary carrying?" A: \textcolor{ForestGreen}{"milk, cup"} & \textbf{Aggregation}. The model must be able to indentify and maintain entity states (e.g., what Mary picked up and didn’t drop). \\
\hline
Indefinite Knowledge & The question is about a fact that the Passage makes ambiguous or uncertain --- answer is "maybe"  or "I don't know" & Passage: "Mary was in school at noon. \{...\} On her way home, Mary visited multiple stores." Q: "Which store did Mary visit last?" A: \textcolor{ForestGreen}{"I don't know"} & \textbf{Refusal (Epistemic Reasoning}). The model must recognise uncertainty, eliminate false options, and refuse over-committing to ambiguous information. \\
\hline
\end{tabular}}
\caption{Description and example of tasks covered in our proposed \dataset\ dataset. }
\label{tab:task-types}
\end{center}
\end{table*}
\newpage

\appendixsection{Distractor Prompts}
\begin{figure}[ht]
\includegraphics[width=\textwidth,keepaspectratio,trim={0 0 0 0},clip]{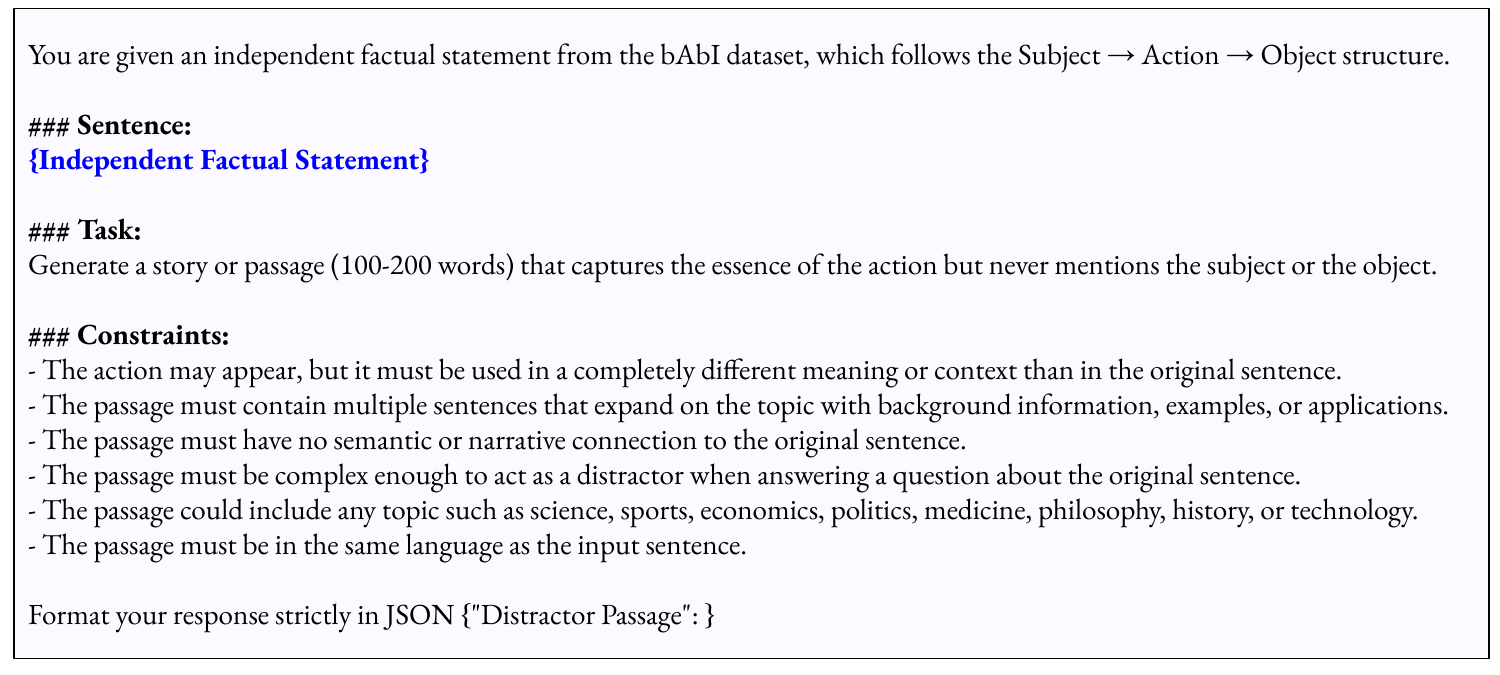}
\caption{Prompt template for generating distractors based on an independent statement (fact) in \dataset, constrained to capture only the action, omitting subject \& object. }
\label{fig:distractor_prompt_1}
\end{figure}

\begin{figure}[ht]
\includegraphics[width=\textwidth,keepaspectratio,trim={0 0 0 0},clip]{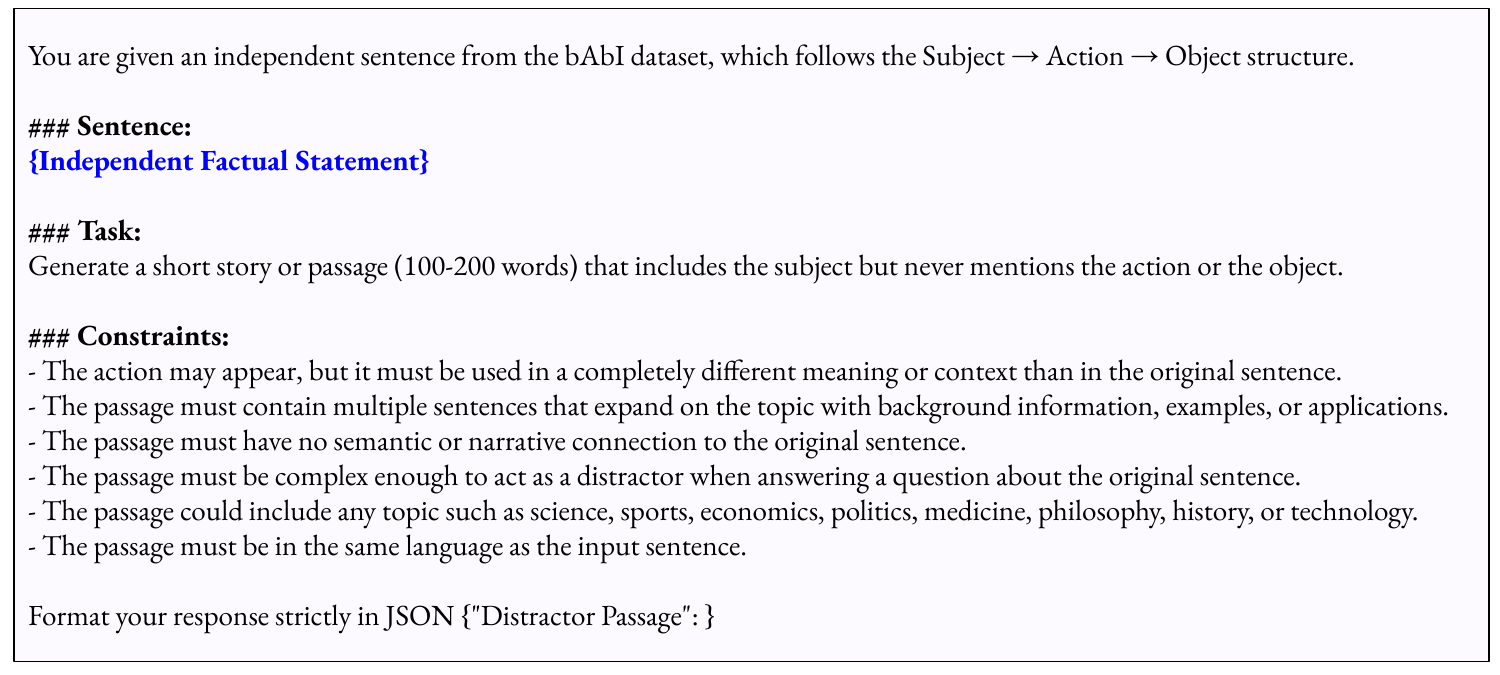}
\caption{Prompt template for generating distractors based on an independent statement (fact) in \dataset,  constrained to include only the object, omitting subject \& action}
\label{fig:distractor_prompt_2}
\end{figure}

\begin{figure}[ht]
\includegraphics[width=\textwidth,keepaspectratio,trim={0 0 0 0},clip]{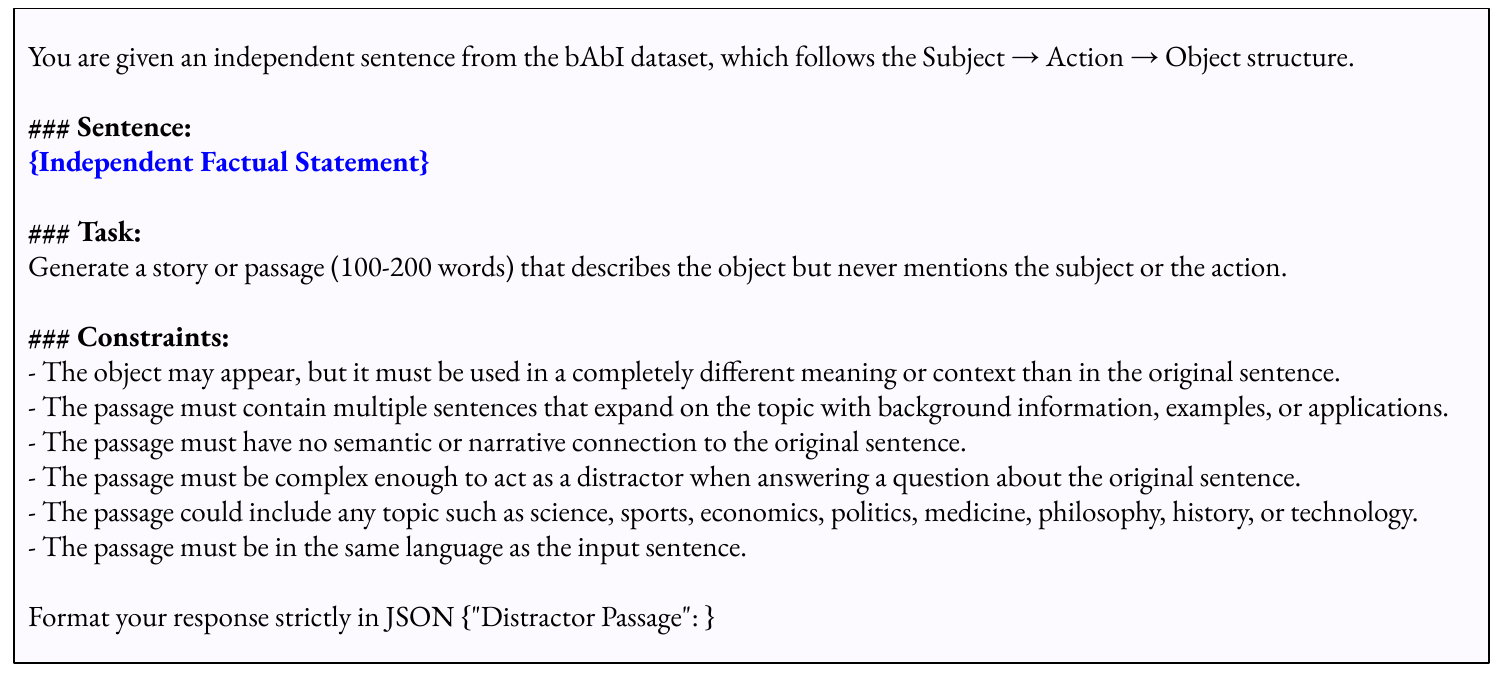}
\caption{Prompt template for generating distractors based on an independent statement (fact) in \dataset. The distractor must capture the action, omitting the subject \& object. }
\label{fig:distractor_prompt_3}
\end{figure}

\begin{figure}[ht]
\includegraphics[width=\textwidth,keepaspectratio,trim={0 0 0 0},clip]{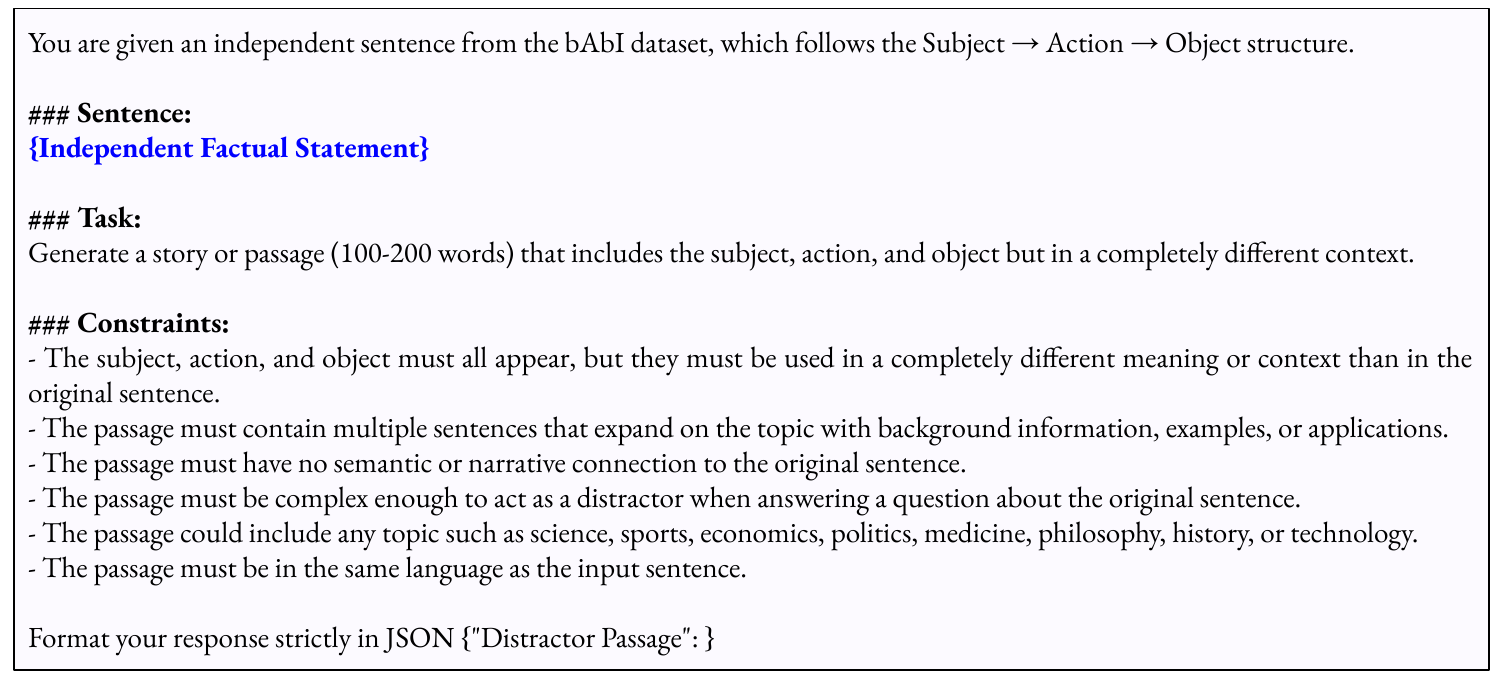}
\caption{Prompt template for generating distractors based on an independent statement (fact) in \dataset, with no constraints on subject, action, and object. }
\label{fig:distractor_prompt_4}
\end{figure}
\newpage
\clearpage

\appendixsection{Prompt Templates}
\begin{figure}[ht]
\includegraphics[width=\textwidth,keepaspectratio,trim={0 0 0 0},clip]{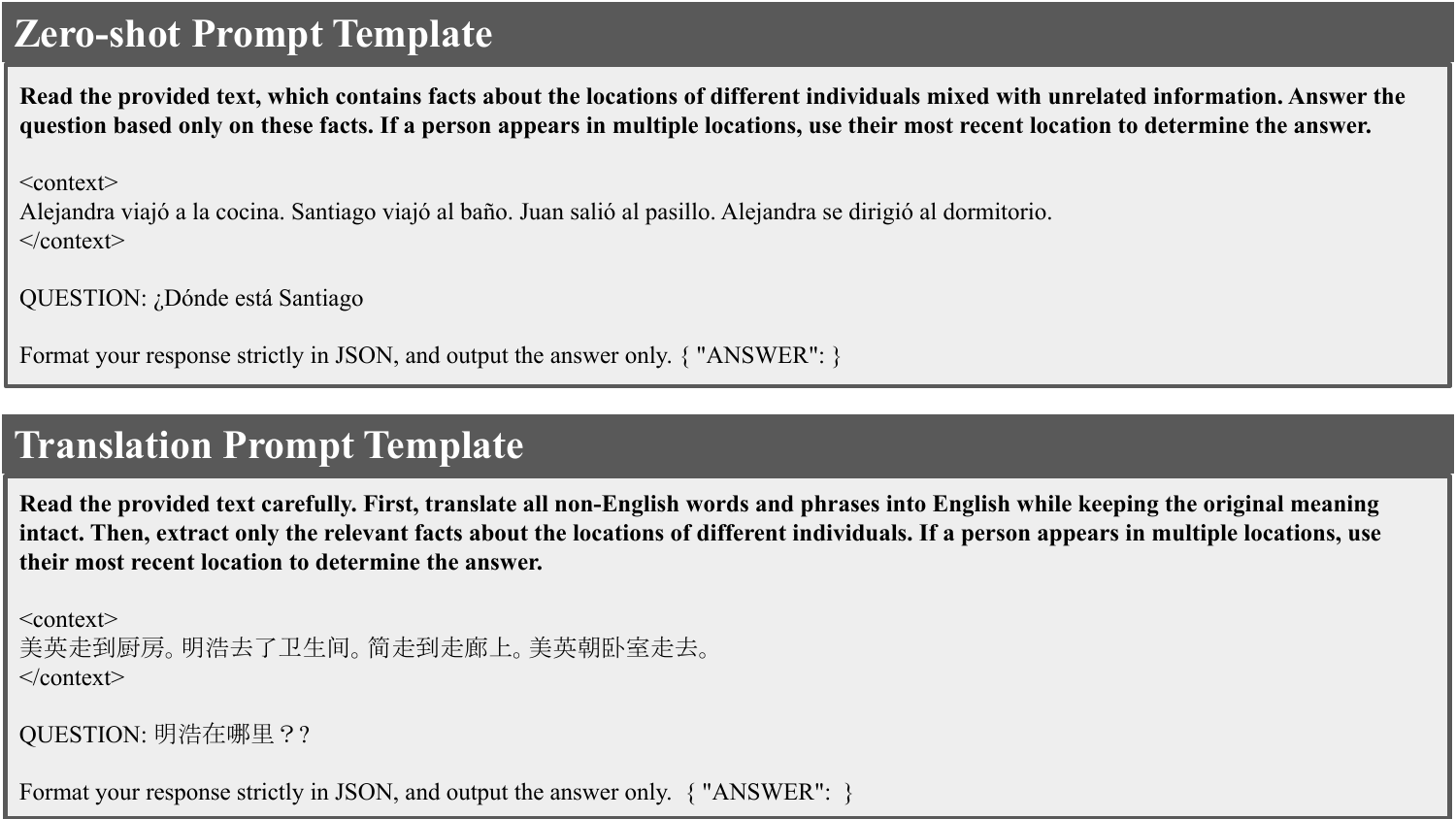}
\caption{Evaluation prompt templates used \textit{zeroshot} and \textit{in-context translation} experiments, respectively.}
\label{fig:zs_prompt_template}
\end{figure}

\begin{figure}[ht]
\includegraphics[width=\textwidth,keepaspectratio,trim={0 0 0 0},clip]{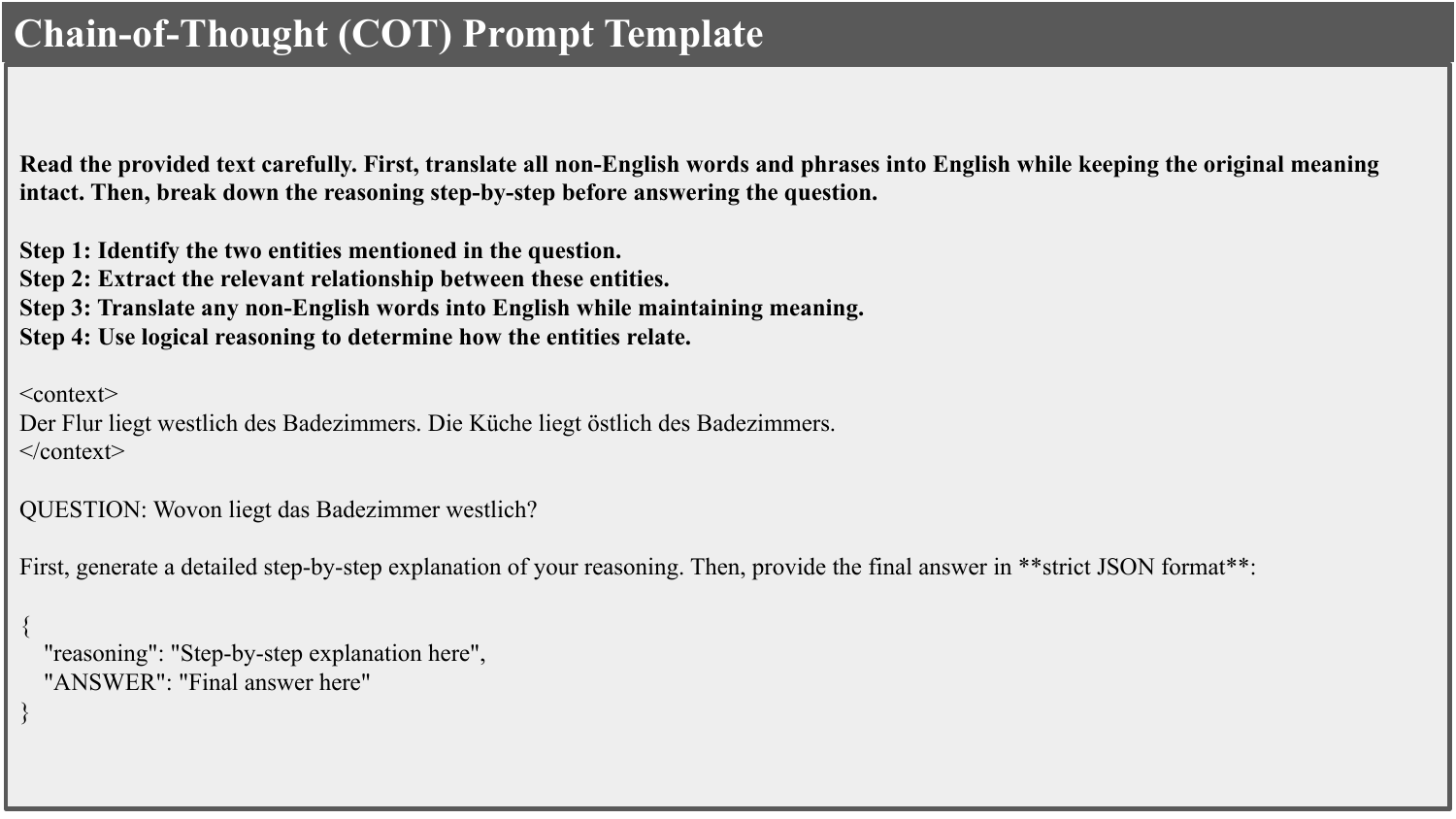}
\caption{Evaluation prompt template used for all \textit{chain-of-thought (CoT)} experiments.}
\label{fig:cot_prompt_template}
\end{figure}

\begin{figure}[ht]
\includegraphics[width=\textwidth,keepaspectratio,trim={0 0 0 0},clip]{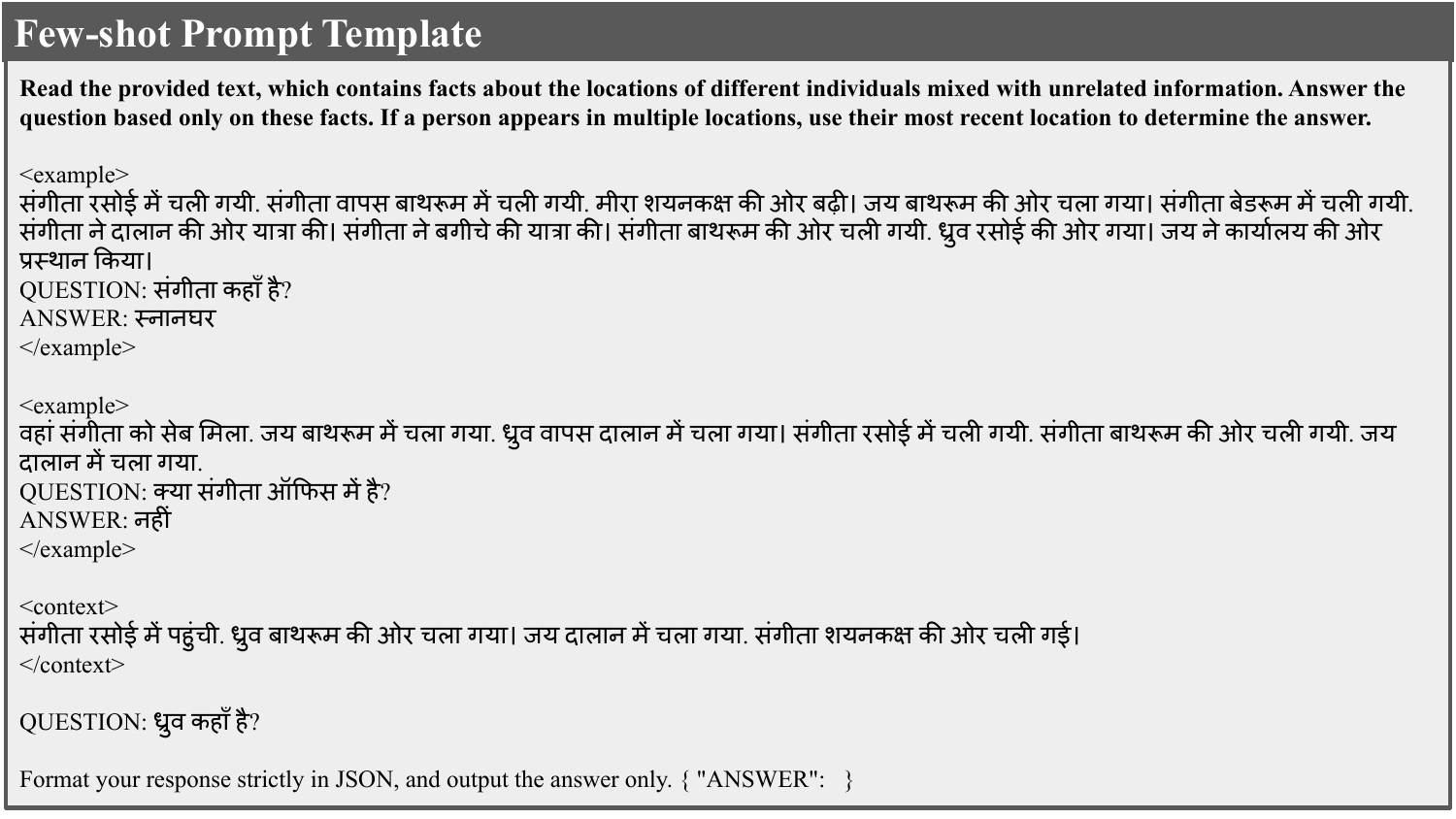}
\caption{Evaluation prompt template used for \textit{fewshot} experiments.}
\label{fig:fs_prompt_template}
\end{figure}
\clearpage
\newpage

\bibliographystyle{compling}
\bibliography{ref}

\end{document}